\documentclass[11pt]{article}
\usepackage{acl}
        
\usepackage{times}
\usepackage{latexsym}

\usepackage[T1]{fontenc}

\usepackage[utf8]{inputenc}

\usepackage{microtype}
\usepackage{inconsolata}

\usepackage{enumitem}
\usepackage{wrapfig}
\usepackage{graphicx}
\usepackage[most]{tcolorbox}
\usepackage{listings}
\usepackage{float}

\tcbset{
  aibox/.style={
    width=206.pt,
    top=1pt,
    bottom=1pt,
    colback=white,
    colframe=lightgray,
    colbacktitle=black,
    enhanced,
    center,
    attach boxed title to top left={yshift=-0.1in,xshift=0.15in},
    boxed title style={boxrule=0pt,colframe=white,},
  }
}
\newtcolorbox{TempBox}[2][]{aibox,title=#2,#1}

\usepackage{multirow}
\usepackage{makecell}
\usepackage{amsmath}
\usepackage[table]{xcolor}
\usepackage{siunitx}
\usepackage{pifont}
\newcommand{\cmark}{{\ding{51}}}

\usepackage{fancyvrb}

\usepackage{subcaption}
\usepackage[framemethod=TikZ]{mdframed}
\usepackage{wrapfig}
\usepackage{amsmath,amsfonts,amssymb}
\usepackage{graphicx}
\usepackage{multirow}
\usepackage{booktabs}
\usepackage{balance}
\usepackage{multicol}
\usepackage{setspace}
\usepackage{pifont}
\usepackage{xcolor}
\usepackage{svg}
\usepackage{dblfloatfix}
\usepackage{enumitem}
\usepackage{lipsum}
\usepackage{tikz}
\usepackage{microtype}
\usepackage[utf8]{inputenc} 
\usepackage[T1]{fontenc}    
\usepackage{hyperref}       
\usepackage{nicefrac}       
\usepackage{microtype}      
\usepackage{lineno}
\usepackage{xspace}
\usepackage{listings}
\usepackage{wrapfig}
\usepackage{bbm}
\usepackage{arydshln}

\usepackage{amsmath,amsfonts,bm}

\def\eqref#1{equation~\ref{#1}}

\def\1{\bm{1}}

\DeclareMathAlphabet{\mathsfit}{\encodingdefault}{\sfdefault}{m}{sl}
\SetMathAlphabet{\mathsfit}{bold}{\encodingdefault}{\sfdefault}{bx}{n}

\usepackage[symbol]{footmisc}

\usepackage[linesnumberedhidden,ruled]{algorithm2e}
\SetAlgoHangIndent{0pt}
\SetKwComment{Comment}{\# }{}
\SetKw{KwInput}{Input:}
\SetKw{KwOutput}{Output:}
\SetKw{KwReturn}{return:}

\makeatletter
\def\adl@drawiv#1#2#3{%
        \hskip.5\tabcolsep
        \xleaders#3{#2.5\@tempdimb #1{1}#2.5\@tempdimb}%
                #2\z@ plus1fil minus1fil\relax
        \hskip.5\tabcolsep}
\newcommand{\cdashlinelr}[1]{%
  \noalign{\vskip\aboverulesep
           \global\let\@dashdrawstore\adl@draw
           \global\let\adl@draw\adl@drawiv}
  \cdashline{#1}
  \noalign{\global\let\adl@draw\@dashdrawstore
           \vskip\belowrulesep}}
\makeatother

\graphicspath{{figures/}}

\usepackage{epigraph} 
\definecolor{tblue}{RGB}{93, 142, 150}

\definecolor{tblue}{RGB}{93, 142, 150}
\definecolor{tred}{RGB}{191, 97, 106}
\definecolor{dlblue}{RGB}{216, 235, 255}
\definecolor{dgreen}{RGB}{74,103,65}
\definecolor{dpink}{RGB}{207, 166, 208}
\definecolor{dyellow}{RGB}{255, 248, 199}
\definecolor{dgray}{RGB}{46, 49, 49}

\newcommand{\durl}[1]{\textcolor{tblue}{\underline{\url{#1}}}}

\newmdenv[
  topline=false,
  bottomline=false,
  rightline = false,
  leftmargin=10pt,
  rightmargin=0pt,
  innertopmargin=0pt,
  innerbottommargin=0pt
]{innerproof}

\newcounter{DaveDefCounter}
\newcommand{\dsr}{DS-R1-distill-Qwen-32B\xspace}
\newcommand{\gametime}{\textsc{GAMETime}\xspace}

\title{Inferring Events from Time Series using Language Models}

\author{
  Mingtian Tan \\
  University of Virginia\\
  \texttt{wtd3gz@virginia.edu} \\
  \And
  Mike A. Merrill \\
  Stanford University \\
  \texttt{mikeam@cs.stanford.edu} \\
  \And
  Zack Gottesman \\
  University of Virginia\\
  \texttt{qdw5jf@virginia.edu} \\
  \AND
  Tim Althoff\\
  University of Washington \\
  \texttt{althoff@cs.washington.edu}
  \And
  David Evans\\
  University of Virginia \\
  \texttt{evans@virginia.edu}
  \And
  Thomas Hartvigsen\\
  University of Virginia \\
  \texttt{hartvigsen@virginia.edu}
}

\begin{document}

\maketitle

\begin{abstract}
A common goal in analyzing time series data is to understand how events cause observed variations.
We study whether Large Language Models (LLMs) can infer natural language events associated with time series data.
We introduce an automated method for generating tasks that test a model's ability to reason about events associated with time series data based on sports data, and develop a new benchmarking method.
In experiments spanning 18 LLMs, we prompt LLMs to infer unobserved events given time series data and observe surprising successes, even when providing minimal context.
We then show that combining distillation with Reinforcement Learning (RL) can improve the performance for small language models to approach that of large proprietary reasoning models.\footnote{All resources needed to reproduce our work are available: \url{https://github.com/hartvigsen-group/GAMETime}}



\end{abstract}

\section{Introduction}{\label{sec:intro}}






\begin{figure*}[t]
    \centering
    \begin{subfigure}{\textwidth}
        \centering
        \includegraphics[width=0.77\linewidth]{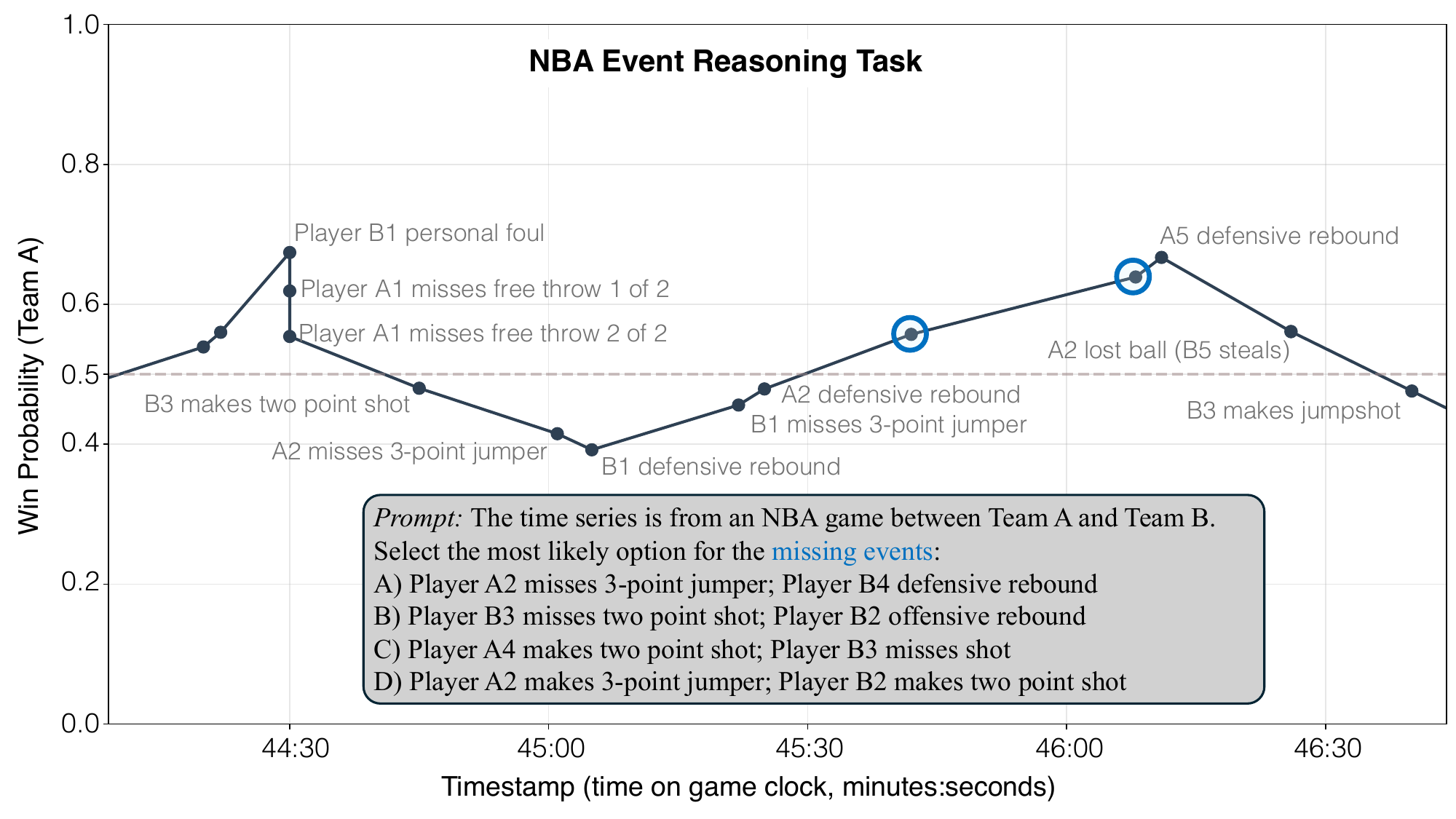}
    \end{subfigure}
    \caption{Illustration of time series event reasoning. The illustrative prompt provides (in text form, see details later in the paper) a time series of win probability values and optional additional context. The model is prompted to select the most likely sequence of events for some segment of the time series data. (This example is derived from an NBA game between the Dallas Mavericks (Team A) and Los Angeles Lakers (Team B), 1 November 2019.) For more details, refer to \autoref{fig:prompt_of_ts_events} in Appendix \ref{app:template}.}
    \label{fig:methods}
\end{figure*}


Time series are time-stamped sequences of values. They are pervasive and come from many sources, like wearable devices ~\cite{anguita2013public}, clinical records ~\cite{harutyunyan2019multitask}, and asset market prices~\cite{wang2024newsforecast,li2024cryptotrade}. 
Although time series analysis typically studies real-valued time series data in isolation, many time series have events that can be described in natural language associated with the numerical data.
Although the causal connections between the numerical data and the natural language events are often uncertain, the events can be correlated with the numerical data and may be causally related. \autoref{fig:methods} illustrates an example using estimated win probabilities for a basketball game---events favorable to Team A increase its win probability, while unfavorable events decrease it. 


Prior work on reasoning about time series in conjunction with natural language has largely overlooked event descriptions~\cite{merrill2024language,williams2024context} and focused on tasks involving just numeric time series data like trend analysis~\cite{cai2024timeseriesexam} or anomaly detection~\cite{dong2024can}. While some studies collect sequences of news related to time series~\cite{wang2024newsforecast,liutime,cheng2024sociodojo}, they are curated for forecasting and do not explore reasoning from the numeric data to events. Meanwhile, due to the potential inclusion of event descriptions that do not impact the time series, as well as failure to include important events, these data are not ideal as a benchmark for measuring LLMs reasoning.







To address this gap, we introduce a method to evaluate a model's ability to reason about the events associated with time series data. The key idea is to measure a model's ability to select the sequence of event descriptions that best correspond to the input time series data. We introduce a benchmark comprising time series data and associated natural language event descriptions.
Our dataset (\autoref{sec3.3:dataset}) includes $4{,}200$ games from the NBA (basketball) and NFL (American football) sports leagues, comprising a total of 1.7 million data points and events.
The real-valued data is \emph{win probability}\footnote{
We use the win probability values output by ESPN's game analysis  (\url{https://www.espn.com/analytics/}). As we discuss in \autoref{sec:limitations}, win probability is an effective measure of game state but is a model's predictions about the future not a measured ground truth value.} and the task, as illustrated in \autoref{fig:events_example}, is to determine which sequence of events is most consistent with the given win probability sequence.


To evaluate the effectiveness of our benchmark in assessing reasoning ability, we test 18 models across various factors, including the impact of available context, varying sequence lengths, and time series similarity on reasoning. We also examine the impact of replacing or removing time series and real entity names through three ablation studies. To explore the generalizability of our approach, we extend the evaluation to open-domain settings, including cryptocurrency prices~\cite{li2024cryptotrade} and U.S. health data~\cite{liutime}. 

Our findings indicate that several LLMs exhibit promising reasoning capabilities. OpenAI's o1 achieved the highest accuracy, predicting 83\% of the NBA events reasoning tasks correctly, followed by \dsr (68\%) and GPT4o (41\%). Through post-training with a distillation phase followed by self-improvement optimization (based on GRPO~\cite{deepseek-math}), we significantly improved the performance of the Qwen2.5 1.5B model from being the worst performing model to outperforming every model except for o1, and approaching its performance on the NBA task.



Our key contribution is a method for programmatically generating tasks that assess an LLMs' ability to reason about events associated with a time series (\autoref{sec3.2:format}). We demonstrate this method using tasks based on sports data and extend it to other domains (\autoref{subsec:other_domains}). We introduce an easily extensible dataset based on continuously-updating sports data with 1.7 million timesteps with values and events (\autoref{sec3.3:dataset}), where changes in time series are explicitly influenced by events. In benchmarking 18 LLMs, we find promising reasoning capabilities and find clear avenues to enhance reasoning (\autoref{subsec:llms} and \autoref{sec:avai_context}) and demonstrate that post-training optimization can significantly enhance LMs ability to infer events associated with a time series (\autoref{sec:post_training}).

\section{Related Work}\label{sec:related_works}


Despite a growing body of work on LLMs and time series reasoning which we summarize in this section, previous benchmarks for LLMs in time series and event reasoning have not addressed the task of inferring event sequences from time series.

\subsection{Time Series Reasoning with LLMs}

\begin{table*}[t]
    \centering
{
    \begin{tabular}{llc@{\hspace{3em}}c@{\hspace{3em}}c}
    \toprule
    & \textbf{Benchmark/Evaluation} & {\bf Context} & {\bf Source} & {\bf Task} \\ 
    \midrule
    \multirow{4}{*}{\rotatebox{90}{Multimodal}} 
    &  \citet{williams2024context} & Description & Manual & Forecasting  \\
    & \citet{merrill2024language} & Description & Synthetic & Reasoning \& Forecasting \\
    & \citet{cai2024timeseriesexam} & Question & Manual & Understanding \\
    & \citet{liutime} & News Series & Real-World & Forecasting \\
    
    \midrule
    \multirow{5}{*}{\rotatebox{90}{Unimodal}} 
    & \citet{fatemi2025test} & Events \& Time & Synthetic & Temporal Reasoning \\
    & \citet{xiong2024large} & Events  \& Time & Synthetic & Temporal Reasoning \\
    & \citet{chu2024timebench} & Events \& Time & Real-World & Temporal Reasoning \\
    & \citet{quan2024econlogicqa} & Events & Synthetic & Sequential Reasoning \\
    & \citet{karger2024forecastbench} & Events & Real-World & Future Forecasting \\
    
    \midrule
    & GAMETime (ours) &  Events \& Time & Real-World & Event Reasoning \\
    \bottomrule
    \end{tabular}
    }
    \caption{Time series benchmarks typically lack a focus on inferring event sequences, while event reasoning evaluations do not incorporate multimodal reasoning over numerical sequences. Our work emphasizes reasoning about events associated with time series data.}
    \label{tab:benchmarks}
\end{table*}

Many studies use text to assist in time series reasoning (including forecasting) with promising results~\cite{cao2023tempo,wang2024chattime,xie2024chatts}. These works have been applied in various domains including sociology~\cite{cheng2024sociodojo}, energy~\cite{wang2024newsforecast,xu2024beyond}, and finance~\cite{li2024cryptotrade,wang2024stocktime}. For example, \citet{williams2024context} manually curated time series forecasting data along with related text to demonstrate the value of incorporating textual information for forecasting tasks. \citet{wang2024newsforecast} used news about energy to help LLMs predict local electricity conditions. Intrinsically, those approaches depend on LLMs' multi-modal transfer of knowledge from natural language to time series. 

However, there are also critical areas where reasoning about real-world events through time series analysis holds significant potential to enhance performance~\cite{jin2024position,jiang2024empowering}, compared to unimodal methods. Using LLMs for anomaly detection~\cite{dong2024can,zhou2024can} often involves processing time series data, such as CPU usage rates from system monitors, and then generating an interpretable anomaly report~\cite{liu2024large}. Similarly, other domains, such as medical care~\cite{chan2024medtsllm}, market analysis~\cite{lee2024quantifying,ye2024beyond}, and human activity analysis~\cite{li2024sensorllm}, also rely on this multi-modal reasoning capability to make actionable decisions.


\autoref{tab:benchmarks} summarizes benchmarks intended to evaluate LLMs' capability in processing time series data. \citet{cai2024timeseriesexam} proposed a benchmark using synthetic data to evaluate LLMs' understanding of time series, focusing on tasks such as pattern recognition. Similarly, ~\citet{merrill2024language} introduced synthetic time series data and relevant textual descriptions, containing a single event (cause), to evaluate LLMs' performance in matching time series to the scenarios that generated them (i.e., etiological reasoning). Due to the lack of paired event sequence, none of these works evaluated the LLMs' ability to reason about events from time series. 

The one exception is \citet{liutime}, which collects news sequences corresponding to time series dating back to 1983. However, due to the limited dataset size and potential contamination issues, it is challenging to use as a fair evaluation source, especially since the exact impact of news on time series remains unclear. To fill this gap, we propose a living benchmark with data sourced from continuously refreshed naturally-occurring data (in our case, from widely available sports data). This avoids the pitfalls associated with synthetic data, and because it can be easily refreshed avoids the contamination risks with fixed benchmarks.


\subsection{Reasoning about Events}
Reasoning is an ill-defined but critical capability, and reasoning benchmarks have been developed for myriad tasks including coding~\cite{zhuo2024bigcodebench,jain2024livecodebench}, mathematics~\cite{cobbe2021training,white2024livebench}, and finance~\cite{xie2023pixiu,islam2023financebench}. Other benchmarks aim to evaluate the general reasoning abilities of LLMs~\cite{bang2023multitask,white2024livebench,srivastava2023beyond}, including  BBH \citep{suzgun2022challenging} and MMLU \citep{hendrycks2020measuring}. 

\autoref{tab:benchmarks} overviews benchmarks that have been proposed to evaluate LLMs' understanding of relationships between events~\cite{quan2024econlogicqa}, as well as temporal reasoning capabilities for understanding the relationships between events and time~\cite{xiong2024large,kenneweg2025traveler}. \citet{karger2024forecastbench} introduced a dynamically updated benchmark to evaluate event forecasting. \citet{fatemi2025test} used synthetic data to assess LLMs' perception and reasoning between events and time. However, these benchmarks do not consider the interplay between time series and associated events, which is the focus of our work. 

\section{Method}\label{sec:methods}

This section describes our method for evaluating reasoning about events associated with a time series. \autoref{sec:problemdefinition} defines the problem setting, \autoref{sec:eventtasks} outlines how we generate tasks from the data, and \autoref{sec:sportsdataset} introduce the dataset we use for our main experiments.




\subsection{Problem Definition}\label{sec:problemdefinition}
A time series is a sequence of timestamped values: $\bm{x} = [(t_0, x_0), (t_1, x_1), ..., (t_T, x_T)]$. The values could be real numbers or textual descriptions of events.
For each sequence, the timestamps $t$ are monotonically increasing ($i < j \implies t_i \leq t_j$), but there can be multiple (ordered) data points with the same timestamp. For our purposes, we consider time series where each data point contains a timestamp along with a real value number and an text event description, although for any particular data point either of these could be missing.

%
%
Although there may not always be a direct causal relationship between the events and measurements, we assume there is some connection between the numeric values and described events.
Note that we do not assume that every timestamp where there is a numeric value includes an event description, or that every timestamp for which there is an event description has a numeric value.


Our goal is to assess an LLM's ability to reason about time series data by measuring its ability to infer unobserved values in a time series $\bm{x}$. As illustrated in \autoref{fig:methods}, when the event descriptions are partially missing, the LLM is tasked to infer the missing events using the provided context and the numeric time series data for the timestamps with missing event descriptions.


\subsection{Generating Event Understanding Tasks}\label{sec3.2:format}\label{sec:eventtasks}
We formulate our event reasoning evaluation as a multiple-choice question~\citep{hendrycks2020measuring,srivastava2023beyond,pal2022medmcqa,white2024livebench} where the model is prompted to select the event descriptions that are most likely to correspond to the provided real-valued time series data. The prompt follows this template:



\begin{TempBox}{}
    System Prompt: 
    \verb|{{sys_prompt}}| \\
    \begin{tabular}{@{\hspace{1em}}l@{\hspace{0.5em}}l@{\hspace{0.5em}}l@{\hspace{0.5em}}l}
    $[t_{i}]$ & $v_{i}$  &  $[s_i]$ & $[e_i]$\\
    $[t_{i+1}]$ & $v_{i+1}$ &  $[s_{i+1}]$ \\
    \multicolumn{2}{c}{$\vdots$} \\   
    $[t_{i+k-2}]$ &  $v_{i+k-2}$  &  $[s_{i+k-2}]$  \\
    $[t_{i+k-1}]$ &  $v_{i+k-1}$ &  $[s_{i+k-1}]$  & $[e_{i+k-1}]$ \\
    \end{tabular} \\
    Four options to choose: \verb|{{options}}|\\
    Respond with this format: \verb|{{format}}|
\end{TempBox}
\label{fig:template}

\noindent where we provide contextual task information (i.e., \verb|sys_prompt|), along with real-valued time series of length $k$, $[(t_i, x_i), \ldots, (t_{i+k-1}, x_{i+k-1})]$. Here, each $x_i$ consists of three components: $v_i$, $s_i$, and $e_i$, representing the win probability, score, and event in our dataset. Some information about some of the intermediate events from $e_{i+1}$ to $e_{i+k-2}$ are missing, and the LLM is tasked with inferring these events. For evaluation, $v_i$ is consistently provided as a baseline for events inferring, while the remaining components are introduced separately as auxiliary conditions. \autoref{fig:prompt_of_nba} in Appendix~\ref{app:template} provides examples of the full prompts used for the baseline setting, where only $v_i$ is provided. To make the task tractable we frame the task as a multiple choice question with four options, one of which corresponds to the actual sequence of events, and prompt the model to select the most likely option. 


To further isolate the LLM's reasoning on time series, we replace specific named entities in our dataset with general, non-identifying descriptors. Specific team names are replaced with \emph{Team A} or \emph{Team B}. Actual player names are replaced with generic labels, such as \emph{Player from Team A}, ensuring that the associations between players and their teams are preserved but revealing no other information about their identities. In evaluations from other domains, such as cryptocurrency prices~\cite{li2024cryptotrade}, we replace all numerical values in news (events) sequence with symbols (e.g., $\alpha$) to prevent LLMs from matching events to time series using dates or price. In open-domain settings, the impact of news on time series may exhibit a minor delay. Therefore, we provide two events occurring before $t_i$ to better capture the full range of events that may influence the time series.

\subsection{A Sports Dataset with Event Descriptions}\label{sec3.3:dataset}\label{sec:sportsdataset}



\begin{figure*}[t]
    \centering
    \begin{subfigure}{\textwidth}
        \centering
        \includegraphics[width=1.0\columnwidth]{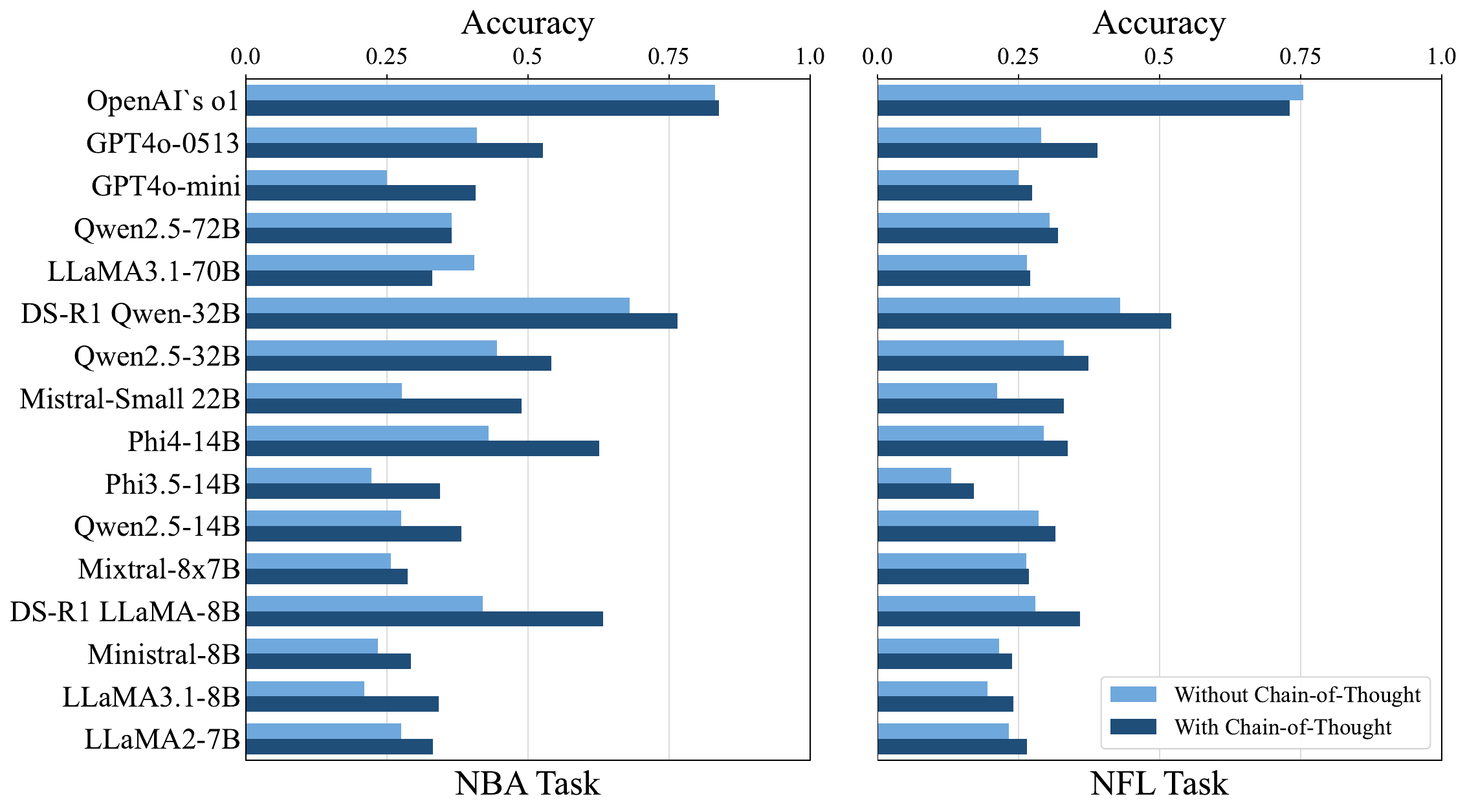}\\
    \end{subfigure}
    \caption{Prediction accuracy on NBA and NFL tasks. 
    The reasoning-focused models such as OpenAI's o1 and DeepSeek-R1 distilled models (DS-R1) significantly outperform others. Chain-of-Thought prompting improves reasoning capabilities for most models. The three proprietary models are first, followed by the open-weight models sorted by decreasing model size.}
    \label{fig:models}
\end{figure*}

To obtain paired data of time series and event sequences, we use data from sports. 
Sports data has two key advantages for our purposes: (1) the in-game events are directly correlated with in-game real-valued data; and (2) the data are continuously refreshed because new games are played every day.
For the natural language events, we used play-by-play data provided by ESPN that describes key events throughout each game, such as scoring, turnovers, or fouls in basketball. 
ESPN also provides each team's predicted win probability throughout the game, which we use as the real-valued time series data. 
These win probabilities reflect the game's status at each time step and changes depending on the events. Since a game is a relatively closed environment, there is a clear relationship between the events and the time series: an event favoring Team A increases Team A’s win probability. This closed environment, along with the continuous generation of new data to avoid contamination problems, makes sports data a good candidate for a benchmark evaluating how effectively LLMs infer events through time series.

We curate \gametime, a new dataset containing $4{,}200$ time series (games) collected through Sep 2025,  $3{,}276$ from NBA basketball games and $924$ from NFL American football games. Examples can be found in Appendix \ref{app:event_example}. Each basketball game contains an average of 460 timesteps, while the football time series average 179 timesteps. The full dataset comprises 1.7 million time data points (win probabilities) paired with corresponding event descriptions.

\section{Experiments}\label{sec:experiments}

To investigate LLMs' event reasoning capabilities under diverse conditions, we explore five research questions: \textbf{RQ1:} Can LLMs reason about events, and does Chain-of-Thought (CoT) prompting enhance this reasoning? \textbf{RQ2:} Can post-training optimization improve event reasoning? \textbf{RQ3:} What is the effect of various available contexts beyond time series? \textbf{RQ4:} Are LLMs able to distinguish underlying time series similarities? \textbf{RQ5:} How do LLMs' event reasoning performance compare across different domains?

\autoref{fig:models} summarizes our results evaluating 18 language models (LMs), including closed-weight models such as GPT-4o \cite{achiam2023gpt} and open-weights models like LLaMA3.1 \cite{dubey2024llama}, and Qwen2.5~\cite{yang2024qwen2}. Models designed for reasoning, including DeepSeek Distilled model, like \dsr~\cite{deepseekr1}, and OpenAI's o1~\cite{openaio1} generally perform better than non-reasoning models. 


\begin{table*}[htb]
\centering
\begin{tabular}{lccrrrrrr}
\toprule
& 
\multicolumn{2}{c}{\bf Post-training} &
\multicolumn{3}{c}{\bf NBA Task} & \multicolumn{3}{c}{\bf NFL Task} \\
\cmidrule(lr){2-3} \cmidrule(lr){4-6} \cmidrule(lr){7-9}
\multicolumn{1}{c}{\bf Model} & \multicolumn{1}{c}{Warmup} & \multicolumn{1}{c}{RL}
& \multicolumn{1}{c}{Correct} & \multicolumn{1}{c}{Incorrect} & \multicolumn{1}{c}{Invalid} & \multicolumn{1}{c}{Correct} & \multicolumn{1}{c}{Incorrect} & \multicolumn{1}{c}{Invalid} \\
\midrule
Qwen2.5 (1.5B) & -- & -- & 11 \hspace{1ex} & 27 \hspace{1ex} &162 \hspace{1ex} &29 \hspace{1ex} &69 \hspace{1ex} &102 \hspace{1ex} \\
Qwen2.5 (1.5B) \hspace{1ex} &\cmark & -- & 111 \hspace{1ex} &82 \hspace{1ex} &7 \hspace{1ex} &69 \hspace{1ex} &128 \hspace{1ex} & 3 \hspace{1ex} \\
Qwen2.5 (1.5B) \hspace{1ex} &-- & \cmark & 32 \hspace{1ex} &114 \hspace{1ex} &54 \hspace{1ex} &43 \hspace{1ex} &111 \hspace{1ex} &46 \hspace{1ex} \\
Qwen2.5 (1.5B) \hspace{1ex} &\cmark & \cmark & 151 \hspace{1ex} &49 \hspace{1ex} &0 \hspace{1ex} &88 \hspace{1ex} &112 \hspace{1ex} &0 \hspace{1ex} \\
GPT4o & -- & -- & 82 \hspace{1ex} &118 \hspace{1ex} &0 \hspace{1ex} &58 \hspace{1ex} &142 \hspace{1ex} &0 \hspace{1ex} \\
DS-R1-32B & -- & -- & 136 \hspace{1ex} &64 \hspace{1ex} &0 \hspace{1ex} &86 \hspace{1ex} &114 \hspace{1ex} &0 \hspace{1ex} \\
OpenAI's o1 & -- & -- & 166 \hspace{1ex} &31 \hspace{1ex} &3 \hspace{1ex} &151 \hspace{1ex} &49 \hspace{1ex} &0 \hspace{1ex} \\
\bottomrule
\end{tabular}
\caption{Comparing the performance of the Qwen2.5 (1.5B) model under different post-training strategies with production models. The numbers are counts of correct, incorrect, and invalid responses from 200 test cases. When applying both warm-up (knowledge distillation) and reinforcement learning (GRPO) post-training, the 1.5B-size model achieves the competitive performance, surpassed only by OpenAI's o1 and outperforming the distillation source \dsr (DS-R1-32B).}
\label{tab:post-training}
\end{table*}






\subsection{Accuracy Evaluation}\label{subsec:llms}
To evaluate LLMs on event inference, we first follow the format in \autoref{fig:prompt_sports} from Appendix \ref{app:template}. In this setting, the model is prompted to select the most likely sequence of events corresponding to a given segment of time series data, where only Team A's win probabilities($\mathit{WP}_A$)\footnote{In NBA basketball there are no draws, and in NFL football draws are exceedingly rare, so the win probability for Team B is $1 - \mathit{WP}_A$.} are provided and the negative options are sequences of the same length randomly sampled from other games. Each model is evaluated on 200 questions. To eliminate memorization effects in reasoning, we select games that occurred after the models’ training cutoff dates and replace real team and player names with generic labels such as \emph{Player from Team A}. 

\autoref{fig:models} summarizes the models' performance on the NBA and NFL task. Although the weakest models barely outperform random guessing, several models, particularly those designed for reasoning, demonstrate strong reasoning performance. GPT-4o achieves an accuracy of $41\%$, and \dsr reaches $68\%$, while o1 performs the best, with an accuracy of $83\%$. Similar results are observed on the NFL data, though the task appears to be more challenging. The performance of GPT-4o drops to $29\%$, while \dsr and o1 achieve $43\%$ and $75.5\%$. Models that failed to generate valid responses, such as Qwen-7B, are not included. Appendix~\ref{app:case_study} shows a case study of how models perform events inferring through time series.


\vspace{1ex}\paragraph{Chain-of-Thought prompting}
Next, we investigate if a longer reasoning process with Chain-of-Thought (CoT) prompting \cite{wei2022chain} improves results of LLMs on event reasoning. LLMs show an average improvement from CoT prompting of $4.5\%$ for the basketball task and $9.6\%$ for the NFL task. In our CoT prompt, we provide an example with a reasoning process (see  \autoref{fig:CoT_basketball} in the Appendix). The longer reasoning process with CoT, however, also slightly increases the overall likelihood of LLMs failing to return answers in the required format by $0.6\%$. We acknowledge the potential for CoT strategy, but we do not further explore this due to computational constraints.


\subsection{Post-training Improves Reasoning}\label{sec:post_training}
The effectiveness of post-training has been demonstrated in math~\cite{deepseekr1}, coding~\cite{openr1}, or vision~\cite{shen2025vlm} tasks. This improvement can be achieved either through training using data containing distilled reasoning processes~\cite{simple1} or through reinforcement learning, such as GRPO~\cite{deepseek-math}. To improve LMs' performance on the event reasoning tasks, we first warm up the LM with knowledge distilled from \dsr, and subsequently apply GRPO training to the warmed-up model.

\begin{table*}[h]\small
    \setlength{\tabcolsep}{2pt} 
    \centering
    \begin{tabular}{cc@{\hspace{1cm}}cccc@{\hspace{1cm}}cccc}
    \toprule
     & & \multicolumn{4}{c}{\bf Available Context} & \multicolumn{3}{c}{\bf Ablations} & \bf Options \\ 
\cmidrule(lr){3-6} \cmidrule(lr){7-9} \cmidrule(lr){10-10}
& \bf Model & \bf WP Only & \bf WP+Times  & \bf WP+Score   & \bf WP+Event 
&  \bf w/ Name & \bf Remove & \bf Replace &  \bf  Reorder   \\ 
\midrule
\multirow{8}{2.5ex}{\rotatebox[origin=c]{90}{{\bf NBA Task}}} 

& GPT4o (0513) & 41.0\% &  \cellcolor{blue!10}{39.0\%}&   47.5\% &   39.0\% &  \cellcolor{blue!25}{\bf 55.0\%}&  \cellcolor{blue!25}{\bf 28.5\%}&   24.0\% &  \cellcolor{blue!25}{\bf 69.5\%}\\  
& GPT4o (mini) & 25.0\% &   24.5\% &   25.0\% &   26.0\% &   43.5\% &   21.0\% &  \cellcolor{blue!10}{27.5\%}&   39.0\% \\  
& Qwen2.5 (72B) & 36.5\% &  \cellcolor{blue!10}{39.0\%}&   43.5\% &  \cellcolor{blue!10}{39.5\%}&   41.0\% &   24.5\% &  \cellcolor{blue!25}{\bf 30.0\%}&  \cellcolor{blue!10}{66.0\%}\\  
& LLama3.1 (70B) &  40.5\% &   37.0\% &  \cellcolor{blue!10}{50.5\%}&   38.5\% &  \cellcolor{blue!10}{51.0\%}&  \cellcolor{blue!10}{26.5\%}&   26.0\% &   47.5\% \\ 
& Qwen2.5 (32B) & \cellcolor{blue!25}{\bf 44.5\%}&  \cellcolor{blue!25}{\bf 43.5\%}&  \cellcolor{blue!25}{\bf 57.5\%}&  \cellcolor{blue!25}{\bf 43.5\%}&   50.0\% &   22.5\% &   26.0\% &   59.0\% \\  
& Phi4 (14B) & \cellcolor{blue!10}{43.0\%}&   35.0\% &   40.0\% &   36.0\% &   42.5\% &   25.0\% &   24.0\% &   47.0\% \\ 
& Qwen2.5 (14B) & 27.5\% &   34.5\% &   33.0\% &   32.0\% &   48.0\% &   22.0\% &   22.0\% &   44.5\% \\  

\cmidrule(lr){2-10}
\multicolumn{3}{c}{\textbf{Avg. Impact of the Condition}}& { $\downarrow$  -0.6\%} & { $\uparrow$  14.6\%}  & { $\uparrow$  0.0\%} & {{ $\uparrow$  33.2\%}} & { $\downarrow$  -32.2\%}  & { $\downarrow$  -27.3\%} & {{ $\uparrow$  46.8\%}}  \\
 
\midrule
\multirow{8}{2.5ex}{\rotatebox[origin=c]{90}{{\bf NFL Task}}}
& GPT4o (0513) & 29.0\% &  \cellcolor{blue!25}{\bf 75.5\%}&  \cellcolor{blue!10}{43.5\%}&  \cellcolor{blue!25}{\bf 53.0\%}&  \cellcolor{blue!25}{\bf 71.0\%}&   18.5\% &   22.0\% &  \cellcolor{blue!25}{\bf 60.0\%}\\  
& GPT4o (mini) & 25.0\% &   52.0\% &   26.5\% &   35.5\% &   33.5\% &   24.5\% &   25.5\% &   42.0\% \\  
& Qwen2.5 (72B) & \cellcolor{blue!10}{30.5\%}&   69.0\% &   42.0\% &   40.5\% &   52.0\% &   25.0\% &   23.0\% &  \cellcolor{blue!10}{54.0\%}\\  
& LLama3.1 (70B) & 26.5\% &   71.0\% &  \cellcolor{blue!25}{\bf 47.5\%}&   35.5\% &  \cellcolor{blue!10}{65.5\%}&   20.5\% &   17.0\% &   46.0\% \\  
& Qwen2.5 (32B) & \cellcolor{blue!25}{\bf 33.0\%}&  \cellcolor{blue!10}{74.5\%}&  \cellcolor{blue!10}{43.5\%}&  \cellcolor{blue!10}{46.0\%}&   40.5\% &  \cellcolor{blue!25}{\bf 27.5\%}&  \cellcolor{blue!25}{\bf 27.0\%}&   43.5\% \\  
& Phi4 (14B) &  29.5\% &   46.5\% &   36.0\% &   38.5\% &   43.5\% &   25.0\% &   23.5\% &   28.5\% \\  
& Qwen2.5 (14B) & 28.5\% &   55.5\% &   28.5\% &   34.5\% &   63.5\% &  \cellcolor{blue!10}{25.5\%}&  \cellcolor{blue!10}{26.0\%}&   33.0\% \\ 

\cmidrule(lr){2-10}
 \multicolumn{3}{c}{\textbf{Avg. Impact of the Condition}} & {{ $\uparrow$  120.1\%}} & { $\uparrow$  32.4\%}  & { $\uparrow$  40.4\%} & {{ $\uparrow$  84.2\%}} & { $\downarrow$  -17.3\%} &  { $\downarrow$  -18.6\%}  & { $\uparrow$  52.8\%}   \\
    
    \bottomrule
    \end{tabular}
    \caption{LLMs' event reasoning accuracy (\%) under various contexts and ablations. We provide each model with 200 questions for each condition. The best performing models are {\bf bold} with \textcolor{blue!40}{blue shading}, and second best are shaded \textcolor{blue!25}{light blue}.}
    \label{tab:avai_context}
\end{table*}

Results in \autoref{tab:post-training} show that even a model with only 1.5B parameters can achieve competitive performance through post-training, surpassing the distilled source and ranking second only to OpenAI o1. For instance, for the NBA task, when we warmed up the model using $3,200$ correctly reasoned Q\&A pairs and their reasoning processes, the warmed-up model correctly infers 111 samples and returns only 7 invalid answers (i.e., no valid result could be extracted) out of 200 test cases, compared to the base model, which answered only 11 correctly and produced 162 invalid results. After further GRPO training with $7,500$ Q\&A pairs, The number of correct reasoning cases reach 151 with no invalid answers, surpassing the distilled model's 136 and approaching the performance of OpenAI's o1. However, employing RL alone without the warm-up phase resulted in only 32 correct responses. Appendix \ref{app:post_train} provide more details on the training process, including reward signals and prompt formats.


\subsection{Impact of Context}\label{sec:avai_context}
In different applications, the available context that LLMs can access varies. Compared to the baseline setting, where only the real-valued time series data is provided, we also evaluate LLMs' performance when different reasoning-relevant contexts are made available or modified. In addition, to evaluate the impact of time series and real names in reasoning and causal relationship between time series and events, we conduct three ablations. The results are summarized in \autoref{tab:avai_context}. 

\vspace{1ex}\paragraph{Available Context}
Due to differences between the football and basketball data, various conditions influence differently. For example, timestamps (\emph{TS+Times}) provide the significant improvement in reasoning for football. Similarly, when providing the score (\emph{TS+Score}) or partial events (\emph{TS+Event}), $e_{i}$ and $e_{i+k-1}$, performance also improves. Note that, given computational constraints and the strong performance of reasoning models, we will primarily focus on avenues to improve base models.


\vspace{1ex}\paragraph{Ablations}
Real player and team names are expected to provide cues that help models identify the correct answer. For example, through potential data contamination or directly matching team names with player names in the options. Results from \emph{w/ Name} column in \autoref{tab:avai_context} demonstrate that real names notably improves accuracy, highlighting the necessity of removing them when evaluating reasoning~\cite{fatemi2025test}. Another two ablations—removing (\emph{Remove}) or replacing (with series from other games) (\emph{Replace}) the time series—model performance drops to near-random levels, indicating that LLMs rely on time series for event inferring and that a strong association exists between the time series and the events.


\vspace{1ex}\paragraph{Options} Due to the nature of possession changes in football and basketball, event sequences follow sequential constraints. To further test whether LLMs can detect logically inconsistent information to aid reasoning, we shuffle the order of ground-truth events to create negative options. Results from the \emph{Reorder} column in \autoref{tab:avai_context} show a clear improvement, indicating that LLMs are capable of leveraging logical sequences through reasoning.

\begin{figure*}[tb]
    \centering
    \begin{subfigure}{0.95\textwidth}
        \centering
        \includegraphics[width=.98\linewidth]{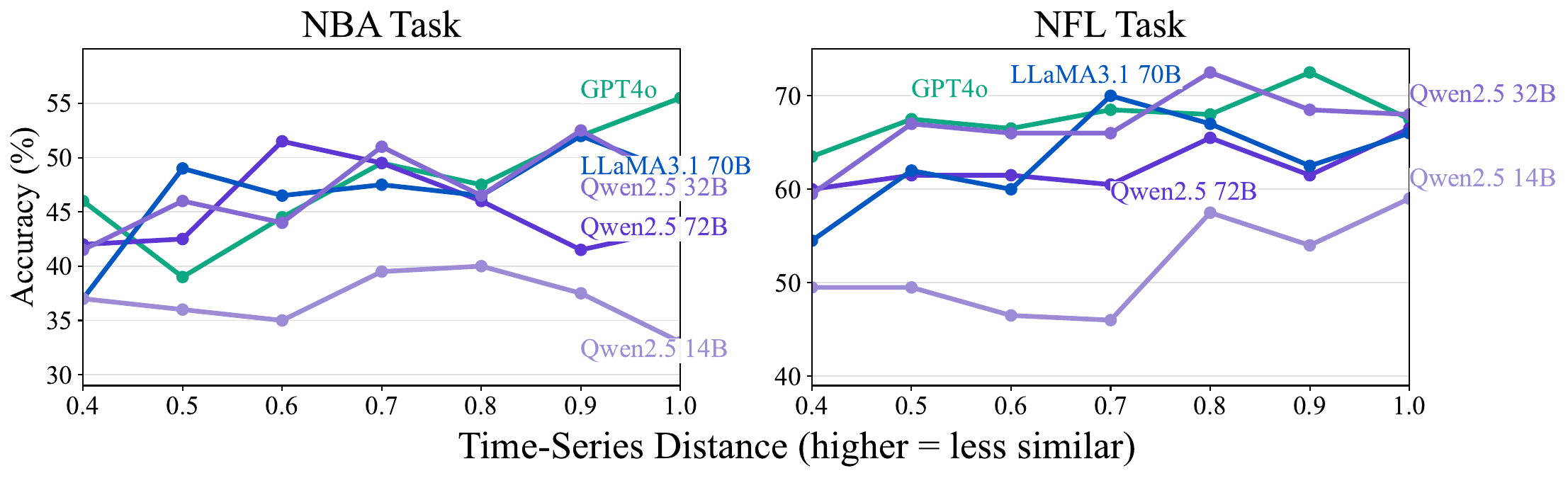}
    \end{subfigure}
    \caption{The performance of LLMs in distinguishing events corresponding to time series (win probabilities) with different levels of similarity. Time series \textit{similarity} decreases as $x$ (i.e., time series \textit{distance}) increases.}
    \label{fig:levels}
\end{figure*}

\subsection{Disparity of Data} 
To assess how the time series similarity impacts LLMs' reasoning, we control the distance between the time series associated with positive and negative options. We compute the distance between time series asEuclidean distance after \emph{z-score normalization}:
$
D=||norm(\mathbf{p}_{win})- norm(\mathbf{p}_{win}^{'})||_{l2}$. 
We divided the distances into seven levels, based on the distribution of win probability differences (see \autoref{fig:dist} in Appendix \ref{app:similarity} for details), starting from 0.4 with an increment of 0.1 per level.

We follow the setup in \autoref{sec3.2:format}, setting the sequence length to 10 and evaluating each LM on 200 questions, keeping the ground-truth events and question time series consistent. The results are presented in \autoref{fig:levels}, showing a slight upward trend in LLM performance as similarity decreases for most of models. This is due to the inherent consistency between time series and event sequences, which LLMs are able to recognize.


\subsection{Other Domains}\label{subsec:other_domains}
Time series data often coexist with related textual sequences. To evaluate the generalizability of our approach, we extend our evaluation to four other domains: trade (import/export, IMEX), health (influenza rates), and energy (gasoline prices) from Time-MMD~\cite{liutime}, as well as cryptocurrency time series from CryptoTrade~\cite{li2024cryptotrade}. To limit the length of prompts, we use news titles as events for cryptocurrency. We selected the ``factual'' field as the events occurring at each timestamp from Time-MMD. \citet{liutime} extracted these ``factual'' statements from news and reports to describe real-world events (Appendix~\ref{app:other_domains}). Our question follows the format in \autoref{sec3.2:format}, with an event sequence length of 10, corresponding to ten trading days for Bitcoin data or ten weeks of influenza statistics in the U.S. health dataset. 

\begin{table}[tb]
    \centering
    \setlength\extrarowheight{0.1pt}
    \addtolength{\tabcolsep}{0.5pt}
    \small
    { 
        \begin{tabular}{cccccccccc}

& 
\rotatebox[origin=l]{90}{{\bf Filtered}}
& \rotatebox[origin=l]{90}{{\bf GPT-4o} (0513)} &
\rotatebox[origin=l]{90}{{\bf GPT-4o} (mini)} &
\rotatebox[origin=l]{90}{{\bf Qwen2.5} (72B)} &
\rotatebox[origin=l]{90}{{\bf DS-R1} (Qwen 32B)} \\

\midrule
        \multirow{2}{*}{\makecell{\textbf{Crypto} \\ (Bitcoin)}}
        & & \cellcolor{blue!25}{\bf 84\%} &  58\% &  \cellcolor{blue!10}{ 71\%} &  62\% &  \\
        & \cmark & \cellcolor{blue!25}{\bf 65\%} & \cellcolor{blue!10}{ 40\%} & \cellcolor{blue!10}{ 40\%} & 39\% & \\
        \midrule
        \multirow{2}{*}{ \makecell{\textbf{Trading} \\ (IMEX)}}
        &  & \cellcolor{blue!10}{ 91\%} &  90\% &  90\% &  \cellcolor{blue!25}{\bf 93\%} &  \\
        & \cmark  & \cellcolor{blue!10}{ 50\%} & 35\% & \cellcolor{blue!25}{\bf 51\%} & 47\% & \\
        \midrule 
        \multirow{2}{*}{\makecell{\textbf{Health} \\ (Influenza)}}
        &  & 62\% &  53\% &  \cellcolor{blue!25}{\bf 77\%} &  \cellcolor{blue!10}{ 74\%} &  \\
        & \cmark  & 33\% & 26\% & \cellcolor{blue!10}{ 34\%} & \cellcolor{blue!25}{\bf 37\%} & \\
        \midrule
        \multirow{2}{*}{\makecell{\textbf{Energy} \\ (Gasoline)}}
        &  & \cellcolor{blue!10}{ 97\%} &  95\% &  96\% &  \cellcolor{blue!25}{\bf 98\%} &  \\
        & \cmark & \cellcolor{blue!25}{\bf 52\%} & 40\% & 48\% & \cellcolor{blue!10}{ 49\%} & \\
        \bottomrule
        \end{tabular}
    }
    \caption{Accuracy of LLMs inferring events across other domains among 100 questions. Replacing numerical information in the events (Filtered rows with \cmark) results in a performance decline compared to retaining the original numbers.} \label{tab:other_domains}
\end{table} 
We evaluate two settings: one where events contain numerical information (i.e., \emph{Complete}) and another where all numerical values, such as dates or real values (e.g., Bitcoin prices or trading volumes), are replaced with symbols like \(\alpha\) (i.e., \emph{Filtered}). \autoref{tab:other_domains} summarizes the results. Even after stripping numerical data, however, LLMs still demonstrate moderate reasoning ability. GPT-4o, for instance, consistently achieves over 50\% accuracy. Additionally, open-weights models such as Qwen2.5 72B or \dsr demonstrate comparable performance to GPT-4o. Detailed results can be found in \autoref{tab:other_domains_full} in Appendix~\ref{app:results}.

\section{Conclusions}
Data comprising time series real values paired with event sequences occur in many important domains. We introduce a dataset containing 1.7 million real-valued time series paired with events and a method for evaluating the ability of an LLM to reason about events corresponding to real-valued time series data. Our evaluation of 18 language models using this benchmark reveals that both open-weights and proprietary models exhibit promising reasoning capabilities, with reasoning models such as \dsr outperforming larger proprietary model such as GPT-4o, while OpenAI's o1 achieves the best performance. By applying post-training optimization, we significantly improve the performance of the Qwen2.5 1.5B to surpassing every model except o1, and approaching o1’s performance on the NBA task.


\section*{Limitations}\label{sec:limitations}
Our dataset includes time series representing win probabilities in sports, which serve as a effective measurement of how events affect a team's state and have a clear relationship with events. Since it is impossible to know the true underlying probability of the game outcome, these probabilities are estimates of each team's chances to win the game produced by ESPN's proprietary model, and not the ground truth. Note that we focus on evaluating the performance of current models rather than exploring how our data can be used for reasoning model training, which we leave for future work.




\section*{Acknowledgements}
This work is supported in part by funds provided by the National Science Foundation, Department of Homeland Security, and IBM through the ACTION AI Institute (Award \#2229876) and by Microsoft's Accelerating Foundation Model Research program. TH was supported by CapitalOne and the University of Virginia's National Security Data \& Policy Institute through the U.S. Department of Defense Contracting Activity \#2024-24070100001.
We thank the University of Virginia's High Performance Computing team for providing excellent computing resources.

\bibliography{main}


\appendix
\renewcommand{\thesubsection}{\Alph{subsection}}
\section{Experimental Setup}

\subsection{Datasets and Language models}\label{app:data_llms}
In this section, we introduce NBA and NFL event and time series data through examples from sports datasets. Additionally, we present the models we evaluate and provide details on data from other domains.

\subsubsection{Events and Time Series in Sports}\label{app:event_example}

\begin{figure*}[t]
    \centering
    \begin{subfigure}{0.98\textwidth}
        \centering
        \includegraphics[width=.48\linewidth]{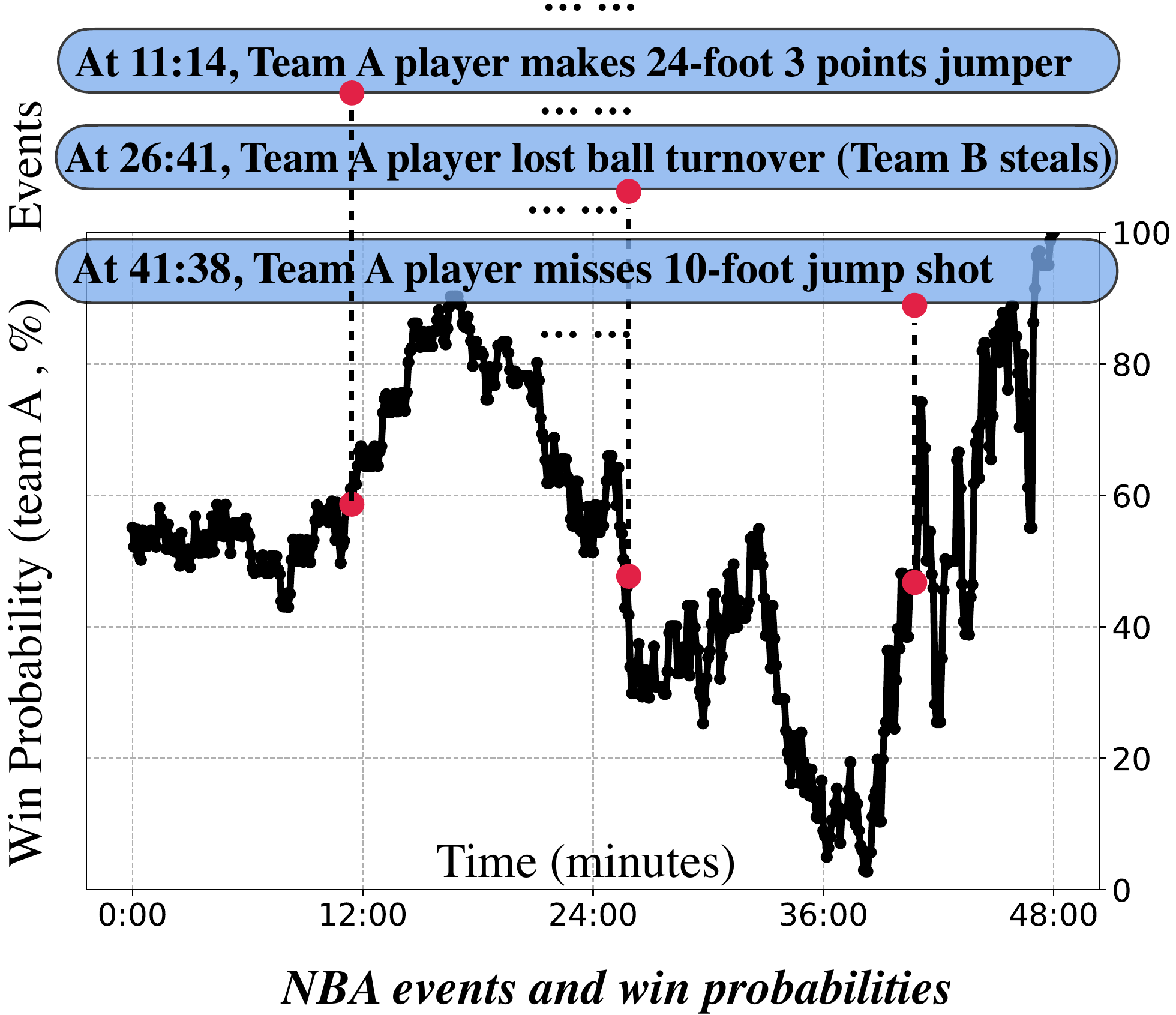}
         \includegraphics[width=.48\linewidth]{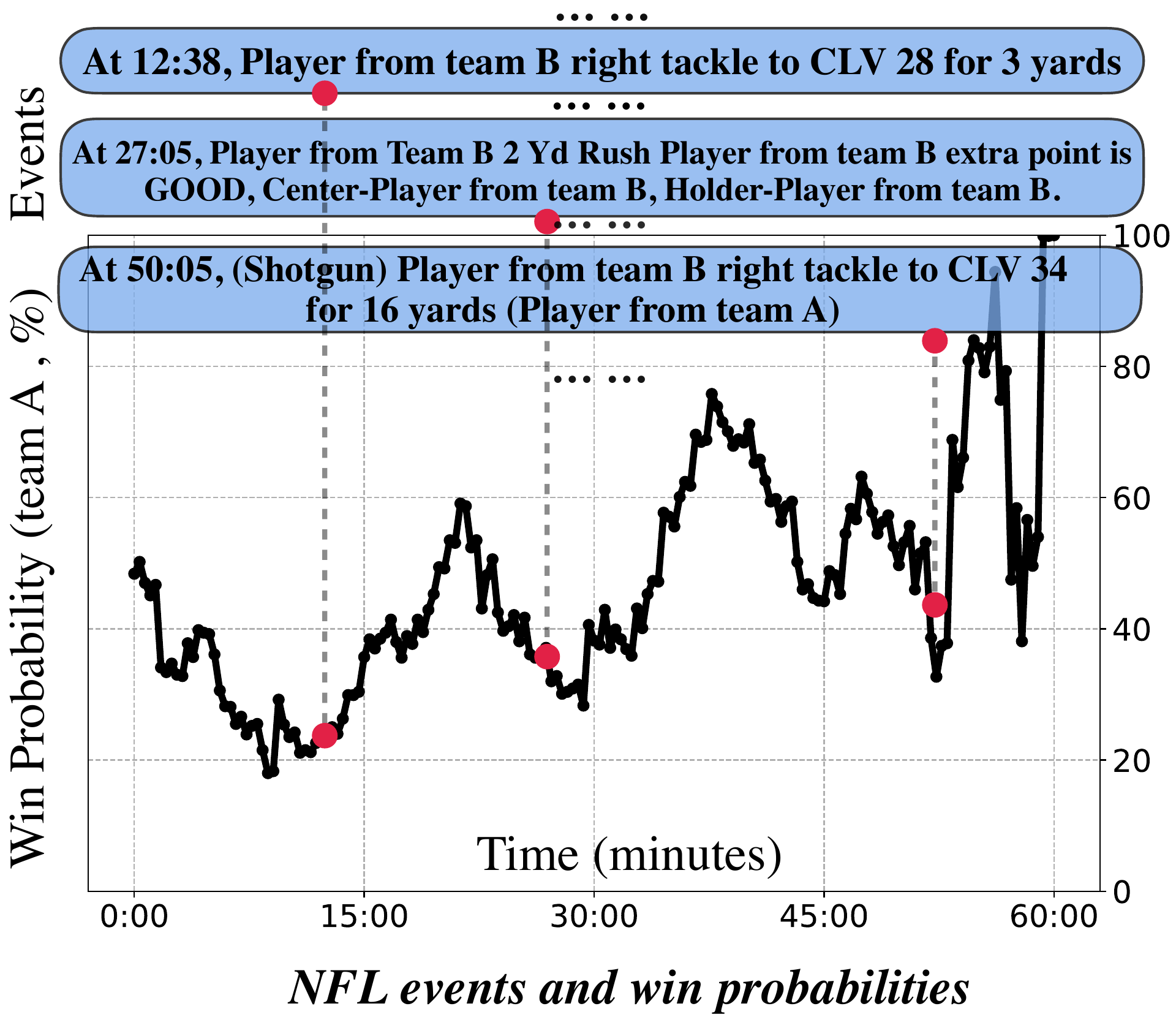}
    \end{subfigure}
    \caption{Examples of events and win probabilities in the NBA and NFL dataset. As the game progresses, ESPN provides descriptions of on-field events along with the corresponding win probabilities for each team at that moment. These probabilities can be considered a representation of the team's current state.}
    \label{fig:events_example}
\end{figure*}

\autoref{fig:events_example} illustrates the time series and event sequences for basketball and football. When an event favorable to Team A occurs, Team A's win probability typically increases. For example, in basketball, this could be a successful score by Team A or a turnover by Team B. In football, it could include defensive plays and sacks by Team A, penalties against Team B, or offensive success by Team A. Conversely, unfavorable events lead to a decrease in win probability.

\subsubsection{Language Models and Setups}\label{app:llms}
We have run our evaluation and experiments on Nvidia A100 GPUs. The specific settings for LLMs, as well as the packages used for data processing, are provided in the repository\footnote{All information and settings needed are available: \url{https://github.com/hartvigsen-group/GAMETime}}. We evaluated a total of 16 models, including open-weight models such as LLaMA3.1~\cite{dubey2024llama}, proprietary models like GPT4o~\cite{achiam2023gpt}, and reasoning-focused models such as DeepSeek-R1~\cite{deepseekr1}. The full list of tested models is as follows:
\begin{itemize}
    \item \textbf{GPT4o}~\cite{achiam2023gpt}: We test GPT4o-0513, a high-performance variant of GPT-4 optimized for both general-purpose generation and specialized tasks, and GPT4o-mini, a scaled-down version of GPT-4 designed for resource-constrained environments.
    \item \textbf{LLaMA}~\cite{dubey2024llama}: We evaluate instruction-tuned models of various parameter sizes, including LLaMA3.1-Instruct 70B, 8B, and LLaMA2-Instruct 7B.
    \item \textbf{Qwen2.5}~\cite{yang2024qwen2}: Our experiments included various instruction-tuned models such as Qwen2.5-Instruct 72B, 32B, 14B, and 8B.
    \item \textbf{Mixtral}~\cite{jiang2024mixtral}: We test the 8x7B Mixture of Experts (MoE) model, along with Mixtral-Small 22B and Ministral-8B.
    \item \textbf{Phi}~\cite{abdin2024phi}: We included Phi-4 14B and Phi-3.5-Instruct 14B in our evaluations.
    \item \textbf{DeepSeek-R1}~\cite{deepseekr1}: Given computational constraints, we still evaluated reasoning-focused models such as DeepSeek-R1 32B and 8B. These models are distilled versions of DeepSeek-R1, using synthetic data from R1 to finetune Qwen 32B and LLaMA 8B, respectively.
\end{itemize}

\subsubsection{Open-world Domains}\label{app:other_domains}
To validate whether LLMs can reason about events through time series in other domains, we utilized four open-world datasets from different fields: Time-MMD~\cite{liutime} (covering Trading, US Health, and Energy) and CryptoTrade~\cite{li2024cryptotrade} (Bitcoin prices). The details are outlined as follows:
\begin{itemize}
    \item \textbf{Trading}: Includes monthly U.S. International Trade Balance data from January 1987 to March 2024 (total length of $423$ months), covering both import and export trade volumes. The corresponding text consists of keyword searches and institutional reports relevant to that month, such as "U.S. International Trade in Goods and Services".
    
    \item \textbf{U.S. Health}: Includes weekly Influenza Patients Proportion data from September 1997 to May 2024 (total length of $1\,389$ weeks). The corresponding text sequences are sourced from weekly keyword searches or reports from the "CDC’s ILINet system".
        
    \item \textbf{Energy}: Contains weekly Gasoline Prices (Dollars per Gallon) from April 1993 to April 2024 (total length of $1\,479$ weeks). The text sequences are obtained through searches or reports from institutions such as the U.S. Energy Information Administration.

    \item \textbf{Bitcoin}: Contains daily Bitcoin price data from January 1, 2023, to February 1, 2024 (time series length of 397), including opening and closing prices, as well as the highest and lowest prices of the day. The corresponding text sequence is derived from authoritative sources such as Bloomberg and Yahoo Finance, filtered through keyword searches to provide five of the most relevant news articles per day. We use their headlines as event descriptions.
\end{itemize}

\subsection{Additional Experimental Results}\label{app:results}

\subsubsection{Other Domains}\label{app:other_domains}
Detailed results of LLMs on other domains are shown in~\autoref{tab:other_domains_full}. It can be observed that LLMs are capable of reasoning about events even in open-world domains. Moreover, when potentially confounding information in the events—such as numbers and dates—is removed (i.e., \emph{Filtered}), LLMs still demonstrate strong reasoning performance.


\begin{table*}[]
    \centering
    \setlength\extrarowheight{0.1pt}
    \addtolength{\tabcolsep}{-2pt}
    \resizebox{\hsize}{!}{ 
        \begin{tabular}{ccccccccccccccccc}
        \toprule
        \textbf{\makecell{LLMs $\rightarrow$ \\ \textbf{ Domains $\downarrow$} }} &  & {\makecell{\textbf{GPT-4o} \\ (0513)}} & {\makecell{\textbf{GPT-4o} \\ (mini)}} & {\makecell{\textbf{Qwen2.5} \\ (72B)}} & {\makecell{\textbf{LLaMA3.1} \\ (70B)}} & {\makecell{\textbf{Mixtral} \\ (8x7B)}} & {\makecell{\textbf{DeepSeek} \\ (R1 32B)}} & {\makecell{\textbf{Qwen2.5} \\ (32B)}} & {\makecell{\textbf{Mistral} \\ (22B)}} & {\makecell{\textbf{Phi4} \\ (14B)}} & {\makecell{\textbf{Qwen2.5} \\ (14B)}} & {\makecell{\textbf{DeepSeek} \\ (R1 8B)}} \\

        \cmidrule(lr){2-13}
        \multirow{2}{*}{\makecell{\textbf{Crypto} \\ (Bitcoin)}}
        &  Complete & \textcolor{red}{84\%} &  58\% &  71\% &  49\% &  36\% &  62\% &  \textcolor{blue}{72\%} &  28\% &  46\% &  51\% &  42\% &  \\
        \cmidrule(lr){3-13}
        &  Filtered & \makecell{\textcolor{red}{65\%} \\ \textcolor{gray}{ $\downarrow$22.6\%}} & \makecell{\textcolor{blue}{40\%} \\ \textcolor{gray}{ $\downarrow$31.0\%}} & \makecell{\textcolor{blue}{40\%} \\ \textcolor{gray}{ $\downarrow$43.7\%}} & \makecell{34\% \\ \textcolor{gray}{ $\downarrow$30.6\%}} & \makecell{29\% \\ \textcolor{gray}{ $\downarrow$19.4\%}} & \makecell{39\% \\ \textcolor{gray}{ $\downarrow$37.1\%}} & \makecell{39\% \\ \textcolor{gray}{ $\downarrow$45.8\%}} & \makecell{27\% \\ \textcolor{gray}{ $\downarrow$3.6\%}} & \makecell{28\% \\ \textcolor{gray}{ $\downarrow$39.1\%}} & \makecell{32\% \\ \textcolor{gray}{ $\downarrow$37.3\%}} & \makecell{25\% \\ \textcolor{gray}{ $\downarrow$40.5\%}} & \\
        
        \cmidrule(lr){2-13}
        \multirow{2}{*}{ \makecell{\textbf{Trading} \\ (IMEX)}}
        &  Complete & \textcolor{blue}{91\%} &  90\% &  90\% &  85\% &  52\% &  \textcolor{red}{93\%} &  86\% &  54\% &  75\% &  71\% &  78\% &  \\
        \cmidrule(lr){3-13}
        &  Filtered & \makecell{\textcolor{blue}{50\%} \\ \textcolor{gray}{ $\downarrow$45.1\%}} & \makecell{35\% \\ \textcolor{gray}{ $\downarrow$61.1\%}} & \makecell{\textcolor{red}{51\%} \\ \textcolor{gray}{ $\downarrow$43.3\%}} & \makecell{36\% \\ \textcolor{gray}{ $\downarrow$57.6\%}} & \makecell{21\% \\ \textcolor{gray}{ $\downarrow$59.6\%}} & \makecell{47\% \\ \textcolor{gray}{ $\downarrow$49.5\%}} & \makecell{45\% \\ \textcolor{gray}{ $\downarrow$47.7\%}} & \makecell{27\% \\ \textcolor{gray}{ $\downarrow$50.0\%}} & \makecell{29\% \\ \textcolor{gray}{ $\downarrow$61.3\%}} & \makecell{31\% \\ \textcolor{gray}{ $\downarrow$56.3\%}} & \makecell{22\% \\ \textcolor{gray}{ $\downarrow$71.8\%}} & \\

        \cmidrule(lr){2-13}
        \multirow{2}{*}{\makecell{\textbf{Health} \\ (Influenza)}}
        &  Complete & 62\% &  53\% &  \textcolor{red}{77\%} &  64\% &  34\% &  \textcolor{blue}{74\%} &  60\% &  24\% &  52\% &  42\% &  48\% &  \\
        \cmidrule(lr){3-13}
        &  Filtered & \makecell{33\% \\ \textcolor{gray}{ $\downarrow$46.8\%}} & \makecell{26\% \\ \textcolor{gray}{ $\downarrow$50.9\%}} & \makecell{\textcolor{blue}{34\%} \\ \textcolor{gray}{ $\downarrow$55.8\%}} & \makecell{27\% \\ \textcolor{gray}{ $\downarrow$57.8\%}} & \makecell{25\% \\ \textcolor{gray}{ $\downarrow$26.5\%}} & \makecell{\textcolor{red}{37\%} \\ \textcolor{gray}{ $\downarrow$50.0\%}} & \makecell{32\% \\ \textcolor{gray}{ $\downarrow$46.7\%}} & \makecell{23\% \\ \textcolor{gray}{ $\downarrow$4.2\%}} & \makecell{30\% \\ \textcolor{gray}{ $\downarrow$42.3\%}} & \makecell{33\% \\ \textcolor{gray}{ $\downarrow$21.4\%}} & \makecell{25\% \\ \textcolor{gray}{ $\downarrow$47.9\%}} & \\
        
        \cmidrule(lr){2-13}
        \multirow{2}{*}{\makecell{\textbf{Energy} \\ (Gasoline)}}
        &  Complete & \textcolor{blue}{97\%} &  95\% &  96\% &  84\% &  63\% &  \textcolor{red}{98\%} &  90\% &  57\% &  89\% &  72\% &  79\% &  \\
        \cmidrule(lr){3-13}
        &  Filtered & \makecell{\textcolor{red}{52\%} \\ \textcolor{gray}{ $\downarrow$46.4\%}} & \makecell{40\% \\ \textcolor{gray}{ $\downarrow$57.9\%}} & \makecell{48\% \\ \textcolor{gray}{ $\downarrow$50.0\%}} & \makecell{46\% \\ \textcolor{gray}{ $\downarrow$45.2\%}} & \makecell{28\% \\ \textcolor{gray}{ $\downarrow$55.6\%}} & \makecell{\textcolor{blue}{49\%} \\ \textcolor{gray}{ $\downarrow$50.0\%}} & \makecell{45\% \\ \textcolor{gray}{ $\downarrow$50.0\%}} & \makecell{24\% \\ \textcolor{gray}{ $\downarrow$57.9\%}} & \makecell{43\% \\ \textcolor{gray}{ $\downarrow$51.7\%}} & \makecell{37\% \\ \textcolor{gray}{ $\downarrow$48.6\%}} & \makecell{29\% \\ \textcolor{gray}{ $\downarrow$63.3\%}} & \\    
        \bottomrule
        \end{tabular}
    }
    \vspace{0mm}
    \caption{The number of correct event reasoning (through time series) made by LLMs across other domains among testing samples (\(N=100\)). Replacing numerical information in the option events—such as dates or prices—with symbols like \(\alpha\) (\emph{Filtered}) results in a performance decline compared to retaining the original numerical information (\emph{Complete}). \textcolor{red}{Red} indicates the best model in this task, while \textcolor{blue}{Blue} represents the second-best.}
    \label{tab:other_domains_full}
\end{table*} 

\subsubsection{Post-training Improves  Inference}\label{app:post_train}
In the post-training phase, we primarily utilize question-answer pairs that included explicit reasoning processes, along with GRPO training, to facilitate the model's self-improvement. The base model employed was Qwen2.5 (1.5B)~\cite{yang2024qwen2}, which demonstrates very limited initial event-inference capabilities. For example, as shown in~\autoref{tab:post-training} for the NFL dataset, it correctly infers 29 out of 200 test cases and produces 102 invalid answers; for the NBA dataset, it correctly reasons only 11 cases and yielded 162 invalid answers.

Inspired by recent work on warming up language models~\cite{simple1,deepseekr1}, we apply knowledge distillation on a relatively strong-performing language model. To avoid data contamination, we selected training data exclusively from games that were different from those used in the test set. Considering the cost and computational resources, we chose \dsr as the distillation source. For the NFL task, we collected a total of 5,434 samples with an accuracy of 44.6\%, and for the NBA task, we collected 4,814 samples with an accuracy of 67.5\%, which is consistent with the results reported in Section \ref{subsec:llms}. We ultimately selected all correctly reasoned samples, along with their reasoning trajectory, to warm up the Qwen (1.5B) model. The prompt structure used for the warm-up is illustrated in Figure \ref{fig:post_training_prompt}. The results (\autoref{tab:post-training}) demonstrate that the warm-up phase significantly improved the model's performance as well as its ability to return valid outputs.

\begin{figure*}[t]
    \centering
    \begin{subfigure}{0.96\textwidth}
        \centering
        \includegraphics[width=\linewidth]{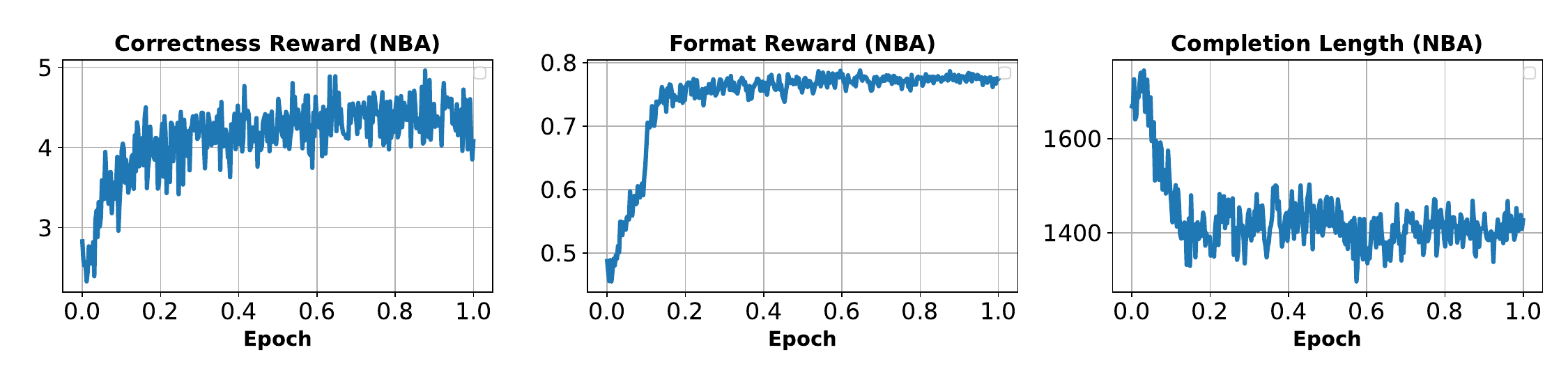} 
        \includegraphics[width=\linewidth]{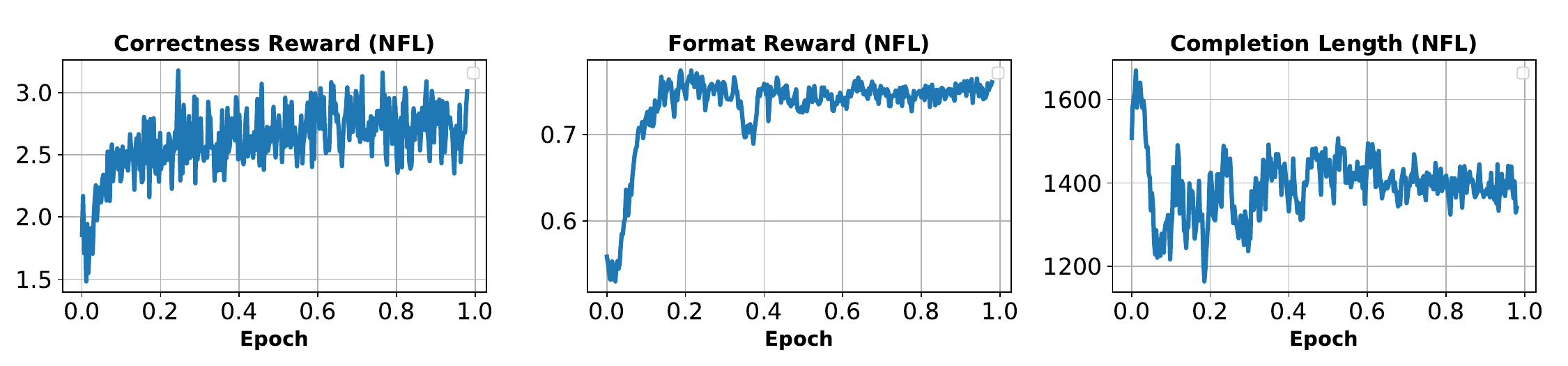}
    \end{subfigure}
    \caption{As the number of training steps increases, both the correctness reward (5 is maximum) and the format reward (0.8 is maximum) show clear improvements, while tokens required to complete the reasoning shows a decreasing trend.}
    \label{fig:reward_steps}
\end{figure*}

Extensive research has shown that self-improvement through optimization leads to significant gains in tasks such as mathematics, coding, and visual reasoning~\cite{shen2025vlm,deepseekr1,deepseek-math}. Building on the warmed-up model, we further applied reinforcement learning using 7,500 Q\&A pairs for each task. The results in \autoref{tab:post-training} show that, after RL optimization (e.g., GRPO~\cite{deepseek-math}), the model surpassed or matched the performance of the distilled model, even though its size was considerably smaller than that of the distillation source. The reasoning template we adopted is shown in \autoref{fig:post_training_prompt}. Specifically, we primarily supervised two types of rewards: \emph{format} and \emph{correctness}, with the training reward trajectories illustrated in~\autoref{fig:reward_steps}. The training was conducted using the open-r1~\cite{openr1} framework and completed on 8 H200 GPUs. Detailed training hyper-parameters and settings are provided in our accompanying repository.

The essence of reinforcement learning in optimizing reasoning is strengthening reasoning trajectory based on reward signals~\cite{liu2503understanding,zhao2025echo,marjanovic2025deepseek}, which requires the language model to possess a certain level of inherent reasoning ability in the task`s domain. Therefore, we also applied GRPO training directly to the base model. Under the same data and training settings, the improvement in performance was limited; however, gains were still observed in the question-answering format, as reflected by a significant reduction in the number of invalid outputs. This further highlights the importance of warming up the model, especially in domains where the base model may have knowledge gaps.

\subsubsection{Number of Events}\label{app:num_of_events}
To further study the effect of event quantity, we follow the setup in \autoref{sec3.2:format} and vary the number of events. Increasing the number of events has two potential effects. On one hand, a competent reasoner should leverage the additional information to identify logical inconsistencies. On the other hand, as the reasoning length increases, the likelihood of errors also rises. A longer reasoning process does not necessarily lead to more accurate results~\cite{wei2022chain}. A capable LLM should ignore any superfluous information and effectively leverage useful context to enhance its reasoning.

\begin{figure}[t]
    \centering
    \begin{subfigure}{0.49\textwidth}
        \centering
        \includegraphics[width=.98\linewidth]{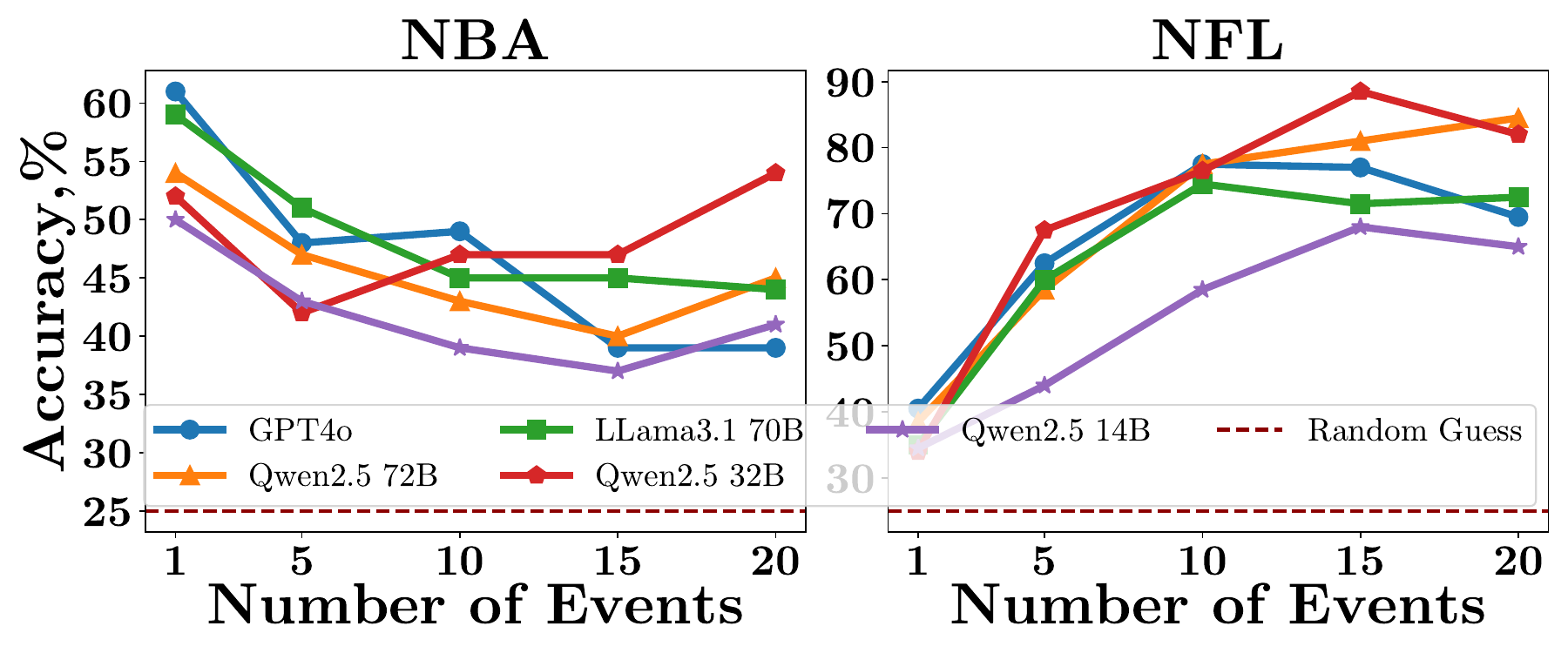}
    \end{subfigure}
    \caption{The reasoning performance of LLMs across event sequences of various lengths. The figure includes only models that consistently outperform the baseline.}
    \label{fig:num_events}
\end{figure}

The results, summarized in \autoref{fig:num_events}, reveal that for the NBA task LLMs generally perform slightly worse as the number of events increases, but for the NFL task performance improves with more events. All else being equal, having more events provides more information and should improve performance; at worst, a strong reasoning model would just ignore additional events and never perform worse. This discrepancy may stem from fundamental differences between the two sports. In a football game, because teams alternate possessions that comprise multiple correlated plays, or events, making it easier to recognize and match patterns. In basketball, each possession is typically connected to only one event, and events are more independent, and most events impact the score of the game directly. In football, each possession involves many events (at least one recorded for each down in football), but most events do not impact the score of the game. \emph{One insight is that the amount of useful information is different across different domains.}

\subsubsection{Time Series Similarity}\label{app:similarity}
\begin{figure}[t]
    \centering
    \begin{subfigure}{0.49\textwidth}
        \centering
        \includegraphics[width=.98\linewidth]{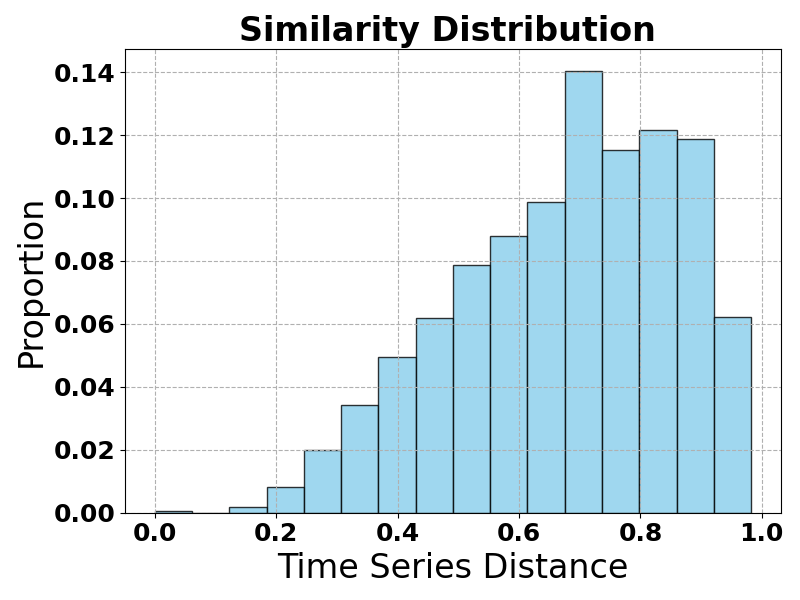}
    \end{subfigure}
    \caption{The similarity distribution of time series in sports data, with a time series length of 10. There is a 91\% probability that the distance between two time series falls beyond 0.4.}
    \label{fig:dist}
\end{figure}

We bootstrap $10k$ pairwise distances between win probabilities (i.e., $\mathbf{p}_{win}$) of length 10 in our dataset and normalize them to the range $(0,1)$. The results show that a large proportion of time series pairs fall within the $(0.4,1)$ range, e.g., 90.6\% for NFL and 91.3\% for NBA data. Their distribution can be shown in \autoref{fig:dist}.

\begin{figure*}[t]
    \centering
    \begin{subfigure}{0.98\textwidth}
        \centering
        \includegraphics[width=.47\linewidth]{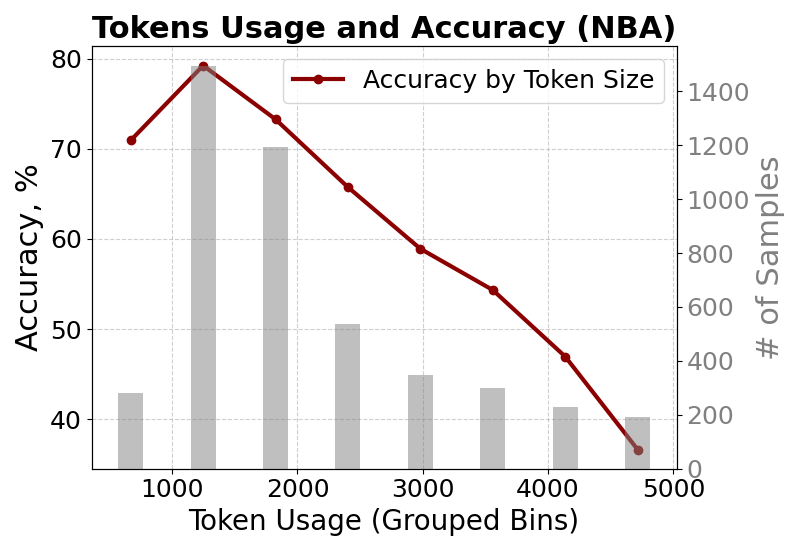}
         \includegraphics[width=.47\linewidth]{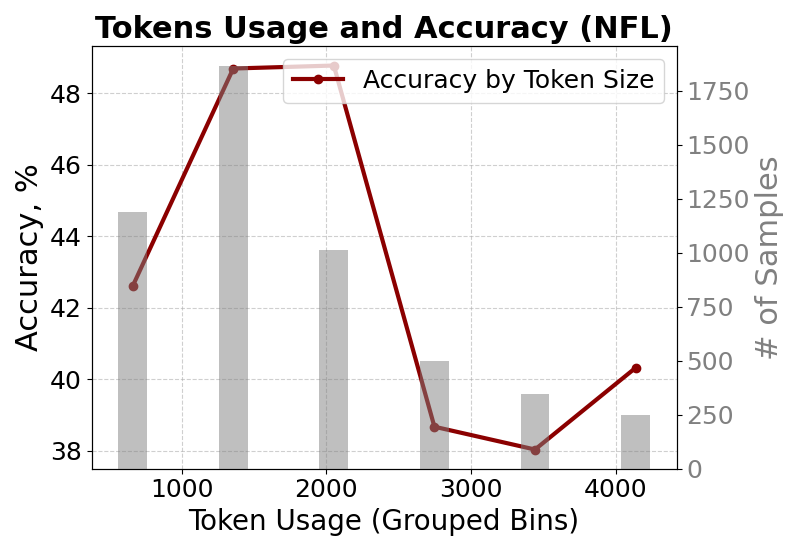}
    \end{subfigure}
    \caption{The relationship between token usage and reasoning accuracy. For both tasks, we sampled around 5,000 examples. We observe that \dsr achieves higher reasoning accuracy when using either fewer or more tokens, with peak accuracy occurring around 1,400 tokens.}
    \label{fig:ds_reasoning_samples}
\end{figure*}

\subsection{Case Study: How LLMs Infer Events}\label{app:case_study}
To further understand how LLMs infer events from time series, we analyze their reasoning process. In this section, we summarize the types of correct and incorrect reasoning process.

\subsubsection{How do language models reason?}\label{app:succ_reasoning}
As shown in \autoref{fig:process}, this illustrates the reasoning process of \dsr~\cite{deepseekr1} for NBA events (under a CoT prompt). The model first interprets the trend in the time series and then matches it with potential events---If the time series exhibits an upward trend, the model aligns it with events favorable to Team A, and vice versa. After sequentially analyzing all data points and their corresponding events, LLMs synthesize their step-by-step analyses to formulate a final reasoning conclusion. High-performing models, such as GPT-4o~\cite{achiam2023gpt}, LLaMA3.1 70B~\cite{dubey2024llama}, Qwen2.5 72B~\cite{yang2024qwen2}, and even smaller language model, like Phi-4~\cite{abdin2024phi} 14B, demonstrate similar reasoning trajectories with CoT Prompting. In addition, for the \dsr, we also observed numerous "aha moments" during the events reasoning process, i.e., self-reflection. For example, in the NBA task, the model reflects midway with \emph{"Wait, maybe the rebound isn't enough".}

\subsubsection{How do language models fail?
}\label{app:failed_reasoning}
We analyzed 5,000 reasoning samples from \dsr, with the results presented in~\autoref{fig:ds_reasoning_samples}. Both excessively short and overly long reasoning processes tend to result in higher error reasoning result. Model accuracy peaks when the reasoning spans approximately 1,400 tokens.

\vspace{1ex}\paragraph{Too Little Reasoning}
We observe that the reasoning errors with short process can largely be attributed to what we term \emph{“rushed reasoning”}. Instead of carefully analyzing each event in the options, as illustrated in~\autoref{fig:ds_reasoning_samples}, the LLM tends to make hasty generalizations and prematurely draws conclusions. An example is shown in~\autoref{fig:case_study_fails}, where the LLM is able to recognize the time series pattern and attempts to reason accordingly. However, it merely provides a superficial summary of each option and arrives at a conclusion after insufficient reasoning.

\vspace{1ex}\paragraph{Too Much Reasoning}
We are not the first to observe that reasoning models, particularly those in the DeepSeek series~\cite{deepseek-math,deepseekr1}, tend to engage in excessively long reasoning when making incorrect inferences~\cite{liu2503understanding,marjanovic2025deepseek}. We categorize these types of errors as cases of \emph{“overthinking,”} characterized by excessive \emph{self-reflection} that leads to confusion and prevents the model from arriving at a correct conclusion. For instance, in~\autoref{fig:ds_reasoning_samples}, case B shows the model repeatedly engaging in self-reflection (e.g., “Wait...”) without reaching a final answer. In this example, the model makes 18 self-corrections. In comparison, the average number of self-reflections in the best-performing range (i.e., token usage between 1,200 to 1,500) is \textbf{7.4}, whereas in “overthinking” cases, where token usage exceeds $3,000$, it rises to an average of \textbf{14.1}.

\begin{figure*}[t]
    \centering
    \begin{subfigure}{0.98\textwidth}
        \centering
        \includegraphics[width=.47\linewidth]{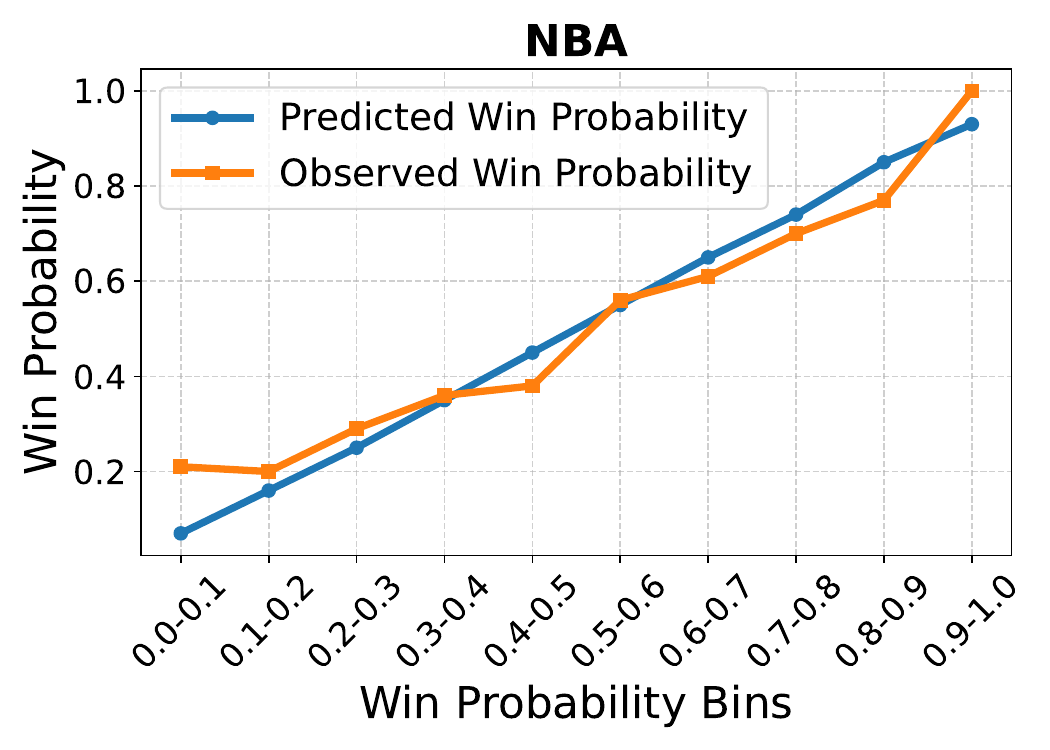}
         \includegraphics[width=.47\linewidth]{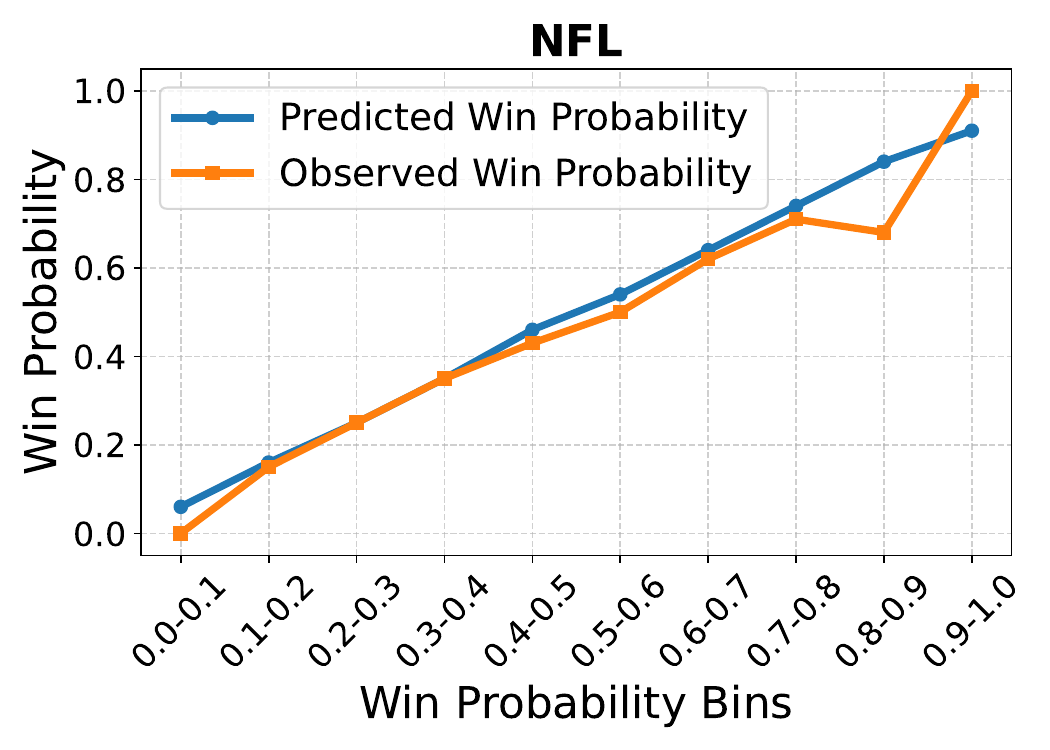}
    \end{subfigure}
    \caption{{The Calibration of Win Probability Predictions.} The results show a high degree of alignment between the model’s predictions and the actual game outcomes.}
    \label{fig:wp_cal}
\end{figure*}

\subsection{Prompt Templates}\label{app:template}
\autoref{fig:prompt_sports} presents the complete template for NBA and U.S Health event reasoning. For NFL data and other domains, we adopt a similar template with minor variations to accommodate domain-specific characteristics. For instance, in cryptocurrency data~\cite{li2024cryptotrade}, we specify that the provided time series represents daily {"Closing Prices,"} while in Energy data~\cite{liutime}, it corresponds to the "Dollars per Gallon." (Gasoline). In addition, considering the delayed impact of real-world news, we included news events from the previous two timestamp in the options. \autoref{fig:prompt_of_nba} and \autoref{fig:prompt_of_ts_events} provide concrete examples of the baseline reasoning setup and the reasoning process with partial events provided, respectively. 

\autoref{fig:CoT_basketball} illustrates the Chain-of-Thought (CoT) prompt for NBA event reasoning, with the format up to the "options" section remaining consistent across prompts. The CoT prompt for NFL follows a similar structure with slight modifications, such as ensuring that example events and background knowledge align with the context of American football. Note that we acknowledge that the current CoT prompt still has room for improvement, however, due to time and computational constraints, we have not conducted further explorations.

\subsection{Win Probability Calibration}\label{app:cal_wp}
To evaluate ESPN’s win probability model, we performed model calibration using the predicted win probabilities at the start of each game and the corresponding outcomes. Specifically, we compared the predicted win rates within each probability bin to the actual win rates observed in those bins, and results in~\autoref{fig:wp_cal} show high degree of consistency between the predictions and the true outcomes.

\lstset{basicstyle=\ttfamily, columns=flexible, breaklines=true, mathescape=true}

\tcbset{
  aibox/.style={
    width=430.pt,
    top=10pt,
    colback=white,
    colframe=lightgray,
    colbacktitle=gray,
    enhanced,
    center,
    attach boxed title to top left={yshift=-0.1in,xshift=0.15in},
    boxed title style={boxrule=0pt,colframe=white,},
  }
}
\newtcolorbox{AIbox}[2][]{aibox,title=#2,#1}

\begin{figure*}[h]
\begin{AIbox}{ Prompt Format of Post-training}
$<|im\_start|>$
Respond reasoning process in the following format:\\
<reasoning>\\
...\\
</reasoning>\\
Return your answer in **X**, where **X** is your answer and can only be one of the selected options, such as **a**, **b**, **c**, or **d**.\\
\verb|{{Question}}| \emph{\qquad\qquad <== The Reinforcement Learning Input Ends Here.} \\
<reasoning> \\ 
\verb|{{Reasoning Process}}| \\ 
</reasoning> \\ 
**\verb|{{Answer}}|**$<|im\_end|>$
\end{AIbox}
\caption{The format of post-training, enclosing the reasoning process within ``\texttt{<reasoning>}'' tags and wrapped the final answer with ``\texttt{**X**}'' to maintain consistency with other evaluation formats.} 
\label{fig:post_training_prompt}
\end{figure*}

\begin{figure*}[h]
\begin{AIbox}{Event Reasoning in Sports (Basketball)}
You are an assistant for NBA basketball task. We will provide a series of consecutive timestamps, win probabilities from a basketball game, though some intermediate events will be missing. You will need to infer the likely events that occurred in the missing intervals.\\
Below is provided timestamps, win probabilities (team A).\\
Step 1. \hspace{0.55cm} TimeStamp$_1$  \hspace{0.55cm}   $WP_1$\\
Step 2. \hspace{0.55cm} TimeStamp$_2$  \hspace{0.55cm}   $WP_2$\\
Step 3. \hspace{0.55cm} TimeStamp$_3$  \hspace{0.55cm}   $WP_3$\\
...\\
Step k. \hspace{0.55cm} TimeStamp$_k$   \hspace{0.55cm} $WP_k$\\
Please select the correct sequence of events for steps $2,...,{k-1}$ from the four options below, \\ 
Here are the potential options:\verb|{{options}}|\\
Here is the instruction for returning reasoning results in:\verb|{{format}}|
\end{AIbox}
\begin{AIbox}{Event Reasoning in Other Domains (U.S Health)}
You are an assistant for an Influenza Patients task. We will provide a series of consecutive timestamps along with the Influenza Patients Proportion. Additionally, we will present four potential event (news) sequences that occurred during that period, as well as from the previous two days. Your task is to identify and select the correct sequence of events.\\
Below is provided date and Patients Proportion (\%),\\
Step 1. \hspace{0.55cm} Date$_0$  \hspace{0.55cm}   $x_0$\\
...\\
Step k. \hspace{0.55cm} Date$_k$   \hspace{0.55cm} $x_k$\\
Please select the correct sequence of events for steps $2,...,{k-1}$ from the four options below, \\ 
Here are the potential options:\verb|{{options}}|\\
Here is the instruction for returning reasoning results in:\verb|{{format}}|
\end{AIbox}
\caption{Template used to infer events through win probabilities in sports (e.g., basketball) and open-world domains (e.g., U.S Health). For NFL data and other domains, we adopt a similar template with minor variations to accommodate domain-specific characteristics. In addition, we use the format "**X**". Time-series timestamps (e.g., dates) can be kept or removed depending on the domain.} 
\label{fig:prompt_sports}
\end{figure*}

\begin{figure*}[h]
\begin{AIbox}{An example of basketball event inferring prompt (Baseline Setting).}
You are an assistant for NBA basketball task. We will provide a series of consecutive win probabilities from a basketball game, though some intermediate events will be missing. You will need to infer the likely events that occurred in the missing intervals.
Below is provided win probabilities (team A).\\
0. 17.1\%\\
1. 16.1\%\\
2. 19.8\%\\
3. 16.6\%\\
4. 15.6\%\\
5. 18.3\%\\
6. 15.3\%\\
7. 10.4\%\\
8. 10.1\%\\
9. 10.1\%\\
10. 10.1\%\\
11. 10.1\%\\
\\
Please select the correct sequence of events for steps 1,2,3,4,5,6,7,8,9,10 from the four options below,\\
a.\\
1.Team A defensive team rebound\\
(... Skip)\\
10.Player from team B enters the game for Player from team B\\
b.\\
1.Player from team A lost ball turnover (Player from team A steals)\\
(... Skip)\\
10.Player from team B misses two point shot\\
c.\\
1.Player from team B enters the game for Player from team B\\
(... Skip)\\
10.Player from team A makes dunk\\
d.\\
1.Player from team A makes two point shot (Player from team A assists)\\
(... Skip)\\
10.Player from team A misses 23-foot three point jumper\\
\\
Return your answer in the format **X**, where X only contains the chosen option, such as **a**, **b**, **c**, or **d**. \\ 
\end{AIbox}
\caption{This illustrates an actual prompt for basketball event reasoning, where the LLM is tasked with inferring the underlying events based solely on a sequence of win probabilities.} 
\label{fig:prompt_of_nba}
\end{figure*}

\begin{figure*}[h]
\begin{AIbox}{An example of event inferring prompt (Time Series and Events are provided).}
You are an assistant for NBA basketball task. We will provide a series of consecutive win probabilities from a basketball game along with a subset of events, though some intermediate events will be missing. You will need to infer the likely events that occurred in the missing intervals. \\ 
Below is provided win probabilities (team A). \\ 
0. 71.1\%  \,\,  Player from team B defensive 3-seconds (technical foul) \\ 
1. 75.9\% \\ 
2. 80.9\% \\ 
3. 81.7\% \\ 
4. 81.7\% \\ 
5. 79.1\% \\ 
6. 76.1\% \\ 
7. 79.2\% \\ 
8. 73.7\% \\ 
9. 83.1\% \\ 
10. 88.0\% \\ 
11. 85.8\% \,\,   Player from team B makes driving layup \\ 
Please select the correct sequence of events for steps 1,2,3,4,5,6,7,8,9,10 from the four options below, \\ 
a. \\ 
1.Player from team B enters the game for Player from team B \\ 
(... Skip) \\ 
10.Player from team B enters the game for Player from team B \\ 
b. \\ 
1.Player from team B defensive rebound \\ 
(... Skip) \\ 
10.Player from team A misses free throw 2 of 2 \\ 
c. \\ 
1.Player from team A makes free throw \\ 
(... Skip) \\ 
10.Player from team A makes 26-foot three point jumper (Player from team A assists) \\ 
d. \\ 
1.Player from team B makes free throw 1 of 2 \\ 
(... Skip) \\ 
10.Player from team A shooting foul \\ 
 \\ 
Please return your answer directly in the format **X**, where X only contains the chosen option, such as **a**, **b**, **c**, or **d**. \\ 

\end{AIbox}
\caption{This illustrates an actual prompt for basketball event reasoning, where the LLM is tasked with inferring the underlying events based on a sequence of win probabilities and the first and the last events during this period.} 
\label{fig:prompt_of_ts_events}
\end{figure*}

\begin{figure*}[h]
\begin{AIbox}{Chain-of-Thought (CoT) Prompt Template for Basketball}
You are an assistant for NBA basketball task ...\\
...

Here are the potential options:\verb|{{options}}|\\

When reasoning, please refer to the following guidelines:\\
1. Analyze Team A's 1st win probability change. For example: \\
   - 0. 53.4\% \\
   - 1. 57.1\% \\
   - Team A's win probability increased from 53.4\% to 57.1\%, indicating that an event positive to Team A or negative to Team B has occurred. \\
2. Pause the win probability analysis for now and focus on the 1st event in options a, b, c, d, and eliminate the inconsistent options or skip this step if unsure. Please ignore very minor changes in win probabilities, such as 0.1\%-0.4\%. Base your decision on significant changes. \\
3. Similarly, turn to analyze the change in the 2nd win probability and match it with the 2nd event in the remaining options, and eliminate the inconsistent options or skip this step if unsure. \\
4. Repeat the above process until all incorrect options are eliminated. \\

Consider the following information when reasoning, it will assist you in reasoning. \\

1. Positive events for Team A, such as "Player from Team A scores (e.g., makes a layup or jump shot) or rebounds," or negative events for Team B, such as "Player from Team B makes a mistake (e.g., misses a shot, commits a foul, or loses the ball)."   \\
2. Some events do not impact the win probability, such as a player entering the game or a timeout. \\
3. Ignore very minor changes in win probabilities, such as 0.1\%-0.4\%.  \\

Return your answer in the format **X**, where X only contains the chosen option, such as **a**, **b**, **c**, or **d**. \\
\end{AIbox}
\caption{Chain-of-Thought (CoT) Template used to infer events through win probabilities in sports (e.g., basketball). For NFL data, we adopt a similar CoT Prompt with minor variations to accommodate domain-specific characteristics.}
\label{fig:CoT_basketball}
\end{figure*}

\begin{figure*}[h]
\begin{AIbox}{Process of Event Reasoning via Time Series From \dsr}
<think> \\ 
Alright, I'm trying to figure out which sequence of events corresponds to the given win probabilities for Team A. The probabilities are as follows:

0. 31.9\%  \\
1. 43.9\%  \\
... ...\\
10. 56.7\%  \\
11. 50.9\%  \\

I need to \textbf{analyze the changes in win probability after each event} and match them with the options provided (a, b, c, d). Let's go through each step one by one. \\

\textbf{Starting with the first change: from 31.9\% to 43.9\%. That's an increase of about 12\%, which is significant. This suggests a positive event for Team A or a negative event for Team B.} Looking at the options, the first events are:\\
a. Player A makes a 3-pointer with an assist.\\
b. Player B makes a free throw 2 of 2.\\
c. Player B defensive rebound.\\
d. Timeout and coach's challenge.\\

\textbf{Option a's first event is a 3-pointer made by Team A, which would increase their win probability.} Option b's first event is a free throw made by Team B, which would decrease Team A's win probability, but the probability went up, so b might be inconsistent.\\
...  \\
So, option a seems the most likely for the first event because scoring a 3-pointer would increase Team A's chances.\\
 ... \\
 
\textbf{Wait, maybe the rebound isn't enough}, ... \\

... \\
The sixth change: from 43.0\% to 49.3\%, an increase of about 6.3\%. Positive event for Team A.\\
....

\textbf{Putting it all together, option a seems to fit all the significant changes in win probability}. The other options don't align as well with the increases and decreases observed.\\
<think> \\
**a**

\end{AIbox}
\caption{Reasoning process example from basketball data: In this case, the ground-truth answer is "a", which aligns with the reasoning of DeepSeek-R1 32B~\cite{deepseekr1}. Other LLMs, such as GPT-4o~\cite{achiam2023gpt}, follow a similar reasoning pattern. To save space, some content has been omitted.} 
\label{fig:process}
\end{figure*}

\subsection{Licensing}\label{app:license}
The code from our work is released under the MIT License, while the dataset is made available under the Creative Commons Attribution-NonCommercial-ShareAlike (CC BY-NC-SA) license. This allows anyone to use, distribute, and modify the data for non-commercial purposes, provided they give proper attribution and share any derivative works under the same license terms. The sports data we use is licensed by the sports leagues to third-party providers like ESPN, which processes the data and adds its win probability model data to it. Because of the complex licensing issues with this data, we do not redistribute it, but our repository includes code for obtaining similar data.

\begin{figure*}[h]
\begin{AIbox}{Case Study A: Failed reasoning with a Short Process}
<think>\\
The win probabilities start at 91.6\% and fluctuate a bit, but overall, Team A's chances seem to be quite high, so I'm guessing the events should reflect a game where Team A is maintaining or slightly increasing their lead. \\ 
Let me look at each option one by one. \\ 
Option a:\\
Team A defensive rebound. That would likely keep their ... \\ 
Option b:\\
Team B offensive rebound. That's bad for Team A because ... \\
Option c:\\
Substitutions happening. That might not directly affect ... \\ 
Option d: \\ 
Team A makes a three-pointer, which is good. Then Team B ... \\ 
I think Option b is better because it shows Team A making baskets and Team B missing, which  ... \\ 
So I'm leaning towards option b. \\ 
</think>\\ 
**b**
\end{AIbox}
\begin{AIbox}{Case Study B: Failed reasoning with a Long Process}
Okay, so I need to figure out which sequence of events (a, b, c, or d) corresponds best to the given win probabilitie ... \\
So, the initial increase from 27.5\% to 29.7\% suggests a positive event for Team A. Then a slight increase again ... \\ 
Starting with option a: \\ 
...\\
Wait, but the first step is from 27.5
...\\
Option b:\\ 
...\\
Wait, but the win probability went up again,\\
...\\
Wait, maybe I'm miscounting. Let me list the events in option b:
...\\
... \\ 
Wait, maybe I'm miscounting the steps. Let me make sure ... \\ 
...\\
Wait, the user provided win probabilities from 0 to 11 ...  \\ 
...\\
Wait, maybe the missed layup is followed by a defensive rebound, ...  \\ 
...\\
Wait, perhaps the win probabilities are for Team A, so when Team B scores, ...  \\ 
...\\
Wait, maybe I miscounted the events. Let me check again. \\ 
...\\
Wait, but the events are listed as steps 1-10, each corresponding ... 
...\\ 
Given all, I think option b is the best fit despite the inconsistency at step 8, because ... \\ 
</think>\\
**b**
\end{AIbox}
\caption{Case A illustrates an example of "rushed reasoning," where the LLM provides only short descriptions of each option before arriving at a conclusion. In contrast, Case B presents an instance of "overthinking," characterized by excessive self-reflection that leads to confusion and ultimately prevents the model from reaching a correct conclusion.} 
\label{fig:case_study_fails}
\end{figure*}

\end{document}